%% file: main_arXiv.tex
\documentclass[10pt,twocolumn,letterpaper]{article}

\usepackage[pagenumbers]{cvpr}      

\input{preamble}

\definecolor{cvprblue}{rgb}{0.21,0.49,0.74}
\usepackage[pagebackref,breaklinks,colorlinks,allcolors=cvprblue]{hyperref}
\usepackage{algorithm}
\usepackage{algorithmic}
\usepackage{booktabs}
\usepackage{multirow}
\usepackage{amsmath}
\usepackage{amssymb}
\usepackage[dvipsnames, table]{xcolor} 
\usepackage{xcolor}
\usepackage{booktabs}

\usepackage{colortbl}
\usepackage{arydshln}

\def\paperID{} 
\def\confName{CVPR}
\def\confYear{2026} 

\title{CompSplat: Compression-aware 3D Gaussian Splatting for Real-world Video}

\author{
Hojun Song$^{*1}$ \quad Heejung Choi$^{*1}$ \quad Aro Kim$^{1}$ \quad Chae-yeong Song$^{1}$ \quad Gahyeon Kim$^{1}$ \\
\quad Soo Ye Kim$^{\dagger 2}$ \quad Jaehyup Lee$^{\dagger 1}$ \quad Sang-hyo Park$^{\dagger 1}$ \\[5pt]
$^{1}$Kyungpook National University \quad $^{2}$Adobe Research \\[6pt]
\begingroup
\hypersetup{urlcolor=cvprblue}
\href{https://hdddhdd.github.io/CompSplat-page/}{\tt\small https://hdddhdd.github.io/CompSplat-page/}
\endgroup
}

\begin{document}
\vspace{-1.0cm}

\twocolumn[{%
  \renewcommand\twocolumn[1][]{#1}%
  \maketitle
  \begin{center}
    \includegraphics[width=\linewidth]{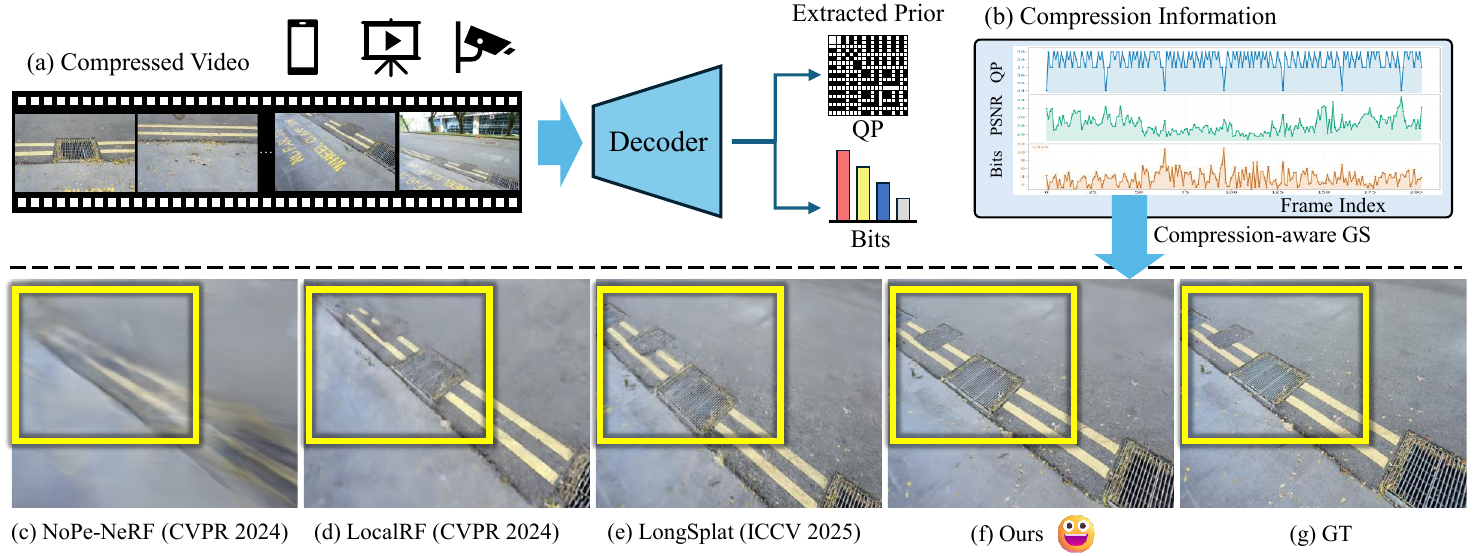}
    \vspace{-0.5cm}
    \captionof{figure}{
      \textbf{CompSplat achieves high-quality novel view synthesis from real-world compressed videos.} Given (a) compressed video input, our approach leverages (b) compression information showing per-frame quality variations from different quantization parameters. Due to degraded inputs from compression, previous methods (c) NoPe-NeRF, (d) LocalRF, and (e) LongSplat generate blurry or distorted results. In contrast, through compression-aware optimization, (f) our proposed method produces clear reconstructions with fine details.
    }
    \label{fig:teaser}
  \end{center}
}]

\maketitle

\let\thefootnote\relax\footnotetext{$^*$Equal contribution. $^\dagger$Corresponding author.}

\input{sec/0_abstract}    
\input{sec/1_introduction}

\input{sec/2_related_work}
\input{sec/3_preliminaries}

\input{sec/4_method}

\input{sec/5_experiments}
\input{sec/6_conclusion}
{
    \small
    \bibliographystyle{ieeenat_fullname}
    \bibliography{main}
}
\input{sec/X_suppl}


\end{document}

%% file: preamble.tex
\usepackage{booktabs, threeparttable, multirow, graphicx}
\usepackage{caption}
\usepackage{makecell}
\newcolumntype{C}{>{\centering\arraybackslash}X}

\usepackage{kotex}
\usepackage{caption}

\usepackage{placeins}

%% file: sec/0_abstract.tex
\begin{abstract}

High-quality novel view synthesis (NVS) from real-world videos is crucial for applications such as cultural heritage preservation, digital twins, and immersive media. However, real-world videos typically contain long sequences with irregular camera trajectories and unknown poses, leading to pose drift, feature misalignment, and geometric distortion during reconstruction. Moreover, lossy compression amplifies these issues by introducing inconsistencies that gradually degrade geometry and rendering quality. While recent studies have addressed either long-sequence NVS or unposed reconstruction, compression-aware approaches still focus on specific artifacts or limited scenarios, leaving diverse compression patterns in long videos insufficiently explored. In this paper, we propose CompSplat, a compression-aware training framework that explicitly models frame-wise compression characteristics to mitigate inter-frame inconsistency and accumulated geometric errors. CompSplat incorporates compression-aware frame weighting and an adaptive pruning strategy to enhance robustness and geometric consistency, particularly under heavy compression. Extensive experiments on challenging benchmarks, including Tanks and Temples, Free, and Hike, demonstrate that CompSplat achieves state-of-the-art rendering quality and pose accuracy, significantly surpassing most recent state-of-the-art NVS approaches under severe compression conditions. 

\end{abstract}

%% file: sec/1_introduction.tex
\section{Introduction}
\label{sec:intro}

High-quality 3D reconstruction and novel view synthesis (NVS) have become central to applications such as virtual and augmented reality, digital twins, and immersive media production \cite{nerf, 3dgs}. By reconstructing reliable three-dimensional structure from captured images and videos, these technologies enable spatial understanding of real scenes and natural human–computer interaction. With the rapid growth of video content on platforms like YouTube and the rising demand for real-world 3D experiences, it is becoming increasingly important to move beyond carefully curated datasets and develop robust methods that remain reliable when operating directly on videos captured in uncontrolled real-world environments \cite{martin2021nerf, wang2022nerf, zhang2024gaussian, shin2025seam360gs}.

Unlike benchmark datasets commonly used in the NVS literature, real-world videos exhibit far more challenging characteristics. Such videos often span hundreds or thousands of frames, contain irregular camera trajectories, and suffer from lighting variations that destabilize pose estimation and geometry recovery \cite{tancik2022block, longsplat}. More importantly, nearly all real-world videos undergo lossy compression, during capture on consumer devices or after upload to online platforms, using codecs such as JPEG~\cite{wallace1991jpeg}, H.264/AVC \cite{wiegand2003overview}, or HEVC \cite{sullivan2012overview}. Compression artifacts from these codecs degrade spatial detail, disrupt temporal coherence, and severely impair downstream processes including keypoint matching, pose estimation, and geometric refinement. 

Existing approaches, which are typically designed and validated on clean and short sequences, struggle to maintain geometric stability or rendering quality when faced with the combined challenges of long-range inconsistency and heterogeneous compression artifacts. The mismatch between controlled research datasets and real-world inputs significantly undermines the practicality of NVS pipelines. Consequently, achieving robust reconstruction from long, compressed video remains as essential yet underexplored challenge. As shown in \cref{fig:teaser}, NoPe-NeRF~\cite{nopenerf}, LocalRF~\cite{localrf}, and LongSplat~\cite{longsplat} produce structure collapse and geometric instability when frames are compressed under typical real-world settings, highlighting the necessity of modeling compression information during reconstruction. 

To address such challenges, we present CompSplat, an adaptive optimization framework tailored for long video sequences under real compression scenarios. CompSplat estimates a frame-wise reliability factor by jointly modeling compression indicators (e.g., quantization parameters, bitrates) and training stability cues including pose estimation confidence and keypoint matching robustness. These weighting factors are continuously updated during training and directly regulate the reconstruction dynamics: frames with high reliability receive denser Gaussian updates, whereas low-quality frames are down-weighted or pruned to prevent the accumulation of compression-induced error and to avoid propagating unstable gradients caused by severely degraded frames. This compression-aware adaptive strategy substantially stabilizes optimization, mitigates geometric distortions, and preserves cross-frame consistency, even in severely compressed or bandwidth-constrained scenarios.

Extensive experiments on compressed versions of long-video datasets, including Free~\cite{free} and Tanks and Temples~\cite{tankandtemples} demonstrate that CompSplat consistently outperforms recent state-of-the-art NVS baselines by large margin. In particular, compared to standard 3DGS methods that do not consider compression, CompSplat significantly improves rendering fidelity, pose accuracy, and overall training stability across diverse compression levels and long-sequence settings. By explicitly modeling real-world degradations, CompSplat meaningfully enhances the practical usability and robustness of Gaussian Splatting-based reconstruction. Our key contributions are summarized as follows:
\begin{itemize}
    \item We propose a compression-aware GS framework (CompSplat) for long compressed real-world videos. 
    \item We firstly design an adaptive frame-wise reliability score that integrates compression metrics and stability cues to dynamically guide Gaussian densification and pruning.
    \item We demonstrate that our CompSplat achieves robust and high-fidelity NVS under practical compression scenarios through comprehensive evaluations across challenging long-sequence GS benchmarks.
\end{itemize}

%% file: sec/2_related_work.tex
\section{Related Work}
\label{sec:Related Work}
\noindent\textbf{Novel View Synthesis.}
Novel view synthesis is a fundamental task in 3D scene understanding that generates images from new viewpoints given a limited number of input images. Neural Radiance Fields (NeRF) \cite{nerf} achieved photorealistic results through implicit neural representations, but it suffers from slow rendering speeds and long optimization times due to volumetric rendering. To overcome these limitations, 3D Gaussian Splatting (3DGS) \cite{3dgs} was proposed. 3DGS represents scenes with millions of Gaussian primitives and enables real-time rendering via differentiable rasterization while maintaining NeRF-level quality. Subsequently, 3DGS research has been conducted in various directions, such as reducing pose estimation dependency \cite{cf3dgs,sfgs}, improving the handling of long videos and large-scale scenes \cite{longsplat, localrf}, and addressing dynamic scene reconstruction \cite{4dgs}. However, most methods were evaluated on high-quality datasets like Mip-NeRF 360 \cite{mipnerf360} and Tanks and Temples \cite{tankandtemples} with uncompressed images, and their robustness on long compressed real-world videos remains underexplored.

\begin{figure*}[t]
    \centering
    \includegraphics[width=\textwidth]{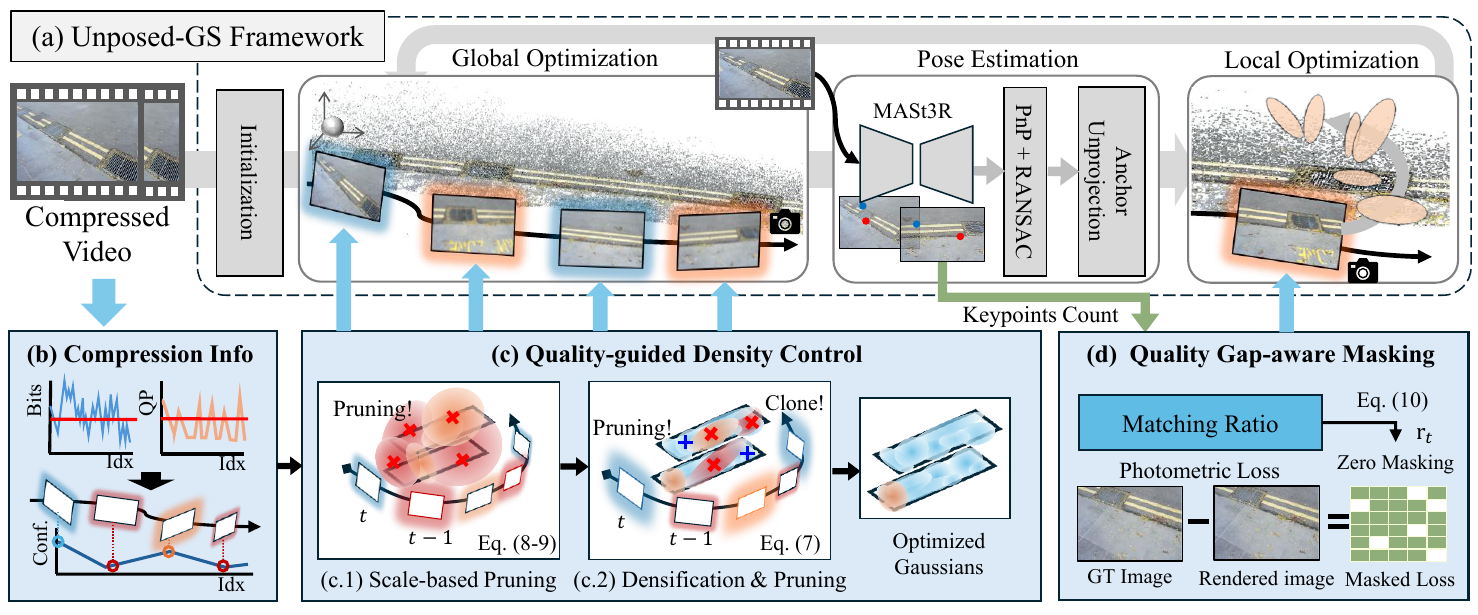}
    \caption{
\textbf{Overview of the CompSplat pipeline: }
(a) Our approach builds upon an unposed-GS framework, reconstructing a 3D Gaussian scene from compressed videos through incremental pose estimation and optimization. 
(b) Frame-wise compression information (QP and bitrates) is converted into a confidence score.
(c) We introduce Quality-guided Density Control, which regulates Gaussian optimization based on frame reliability: 
(c.1) Scale-based pruning removes over-diffused Gaussians that primarily arise in low-quality frames by leveraging frame confidence.  
(c.2) Adaptive Densification and Pruning adjust densification gradient and pruning opacity thresholds based on frame confidence.
(d) Quality Gap-aware Masking mitigates frame-to-frame quality differences by applying a gap ratio–based pixel mask.}
    \label{fig:2}
\end{figure*}

\vspace{0.2em}
\noindent\textbf{3D Reconstruction from Real-World Videos.}
Real-world videos often contain long sequences, irregular camera motion, and incomplete pose information, leading to pose drift, unstable feature matching, and temporal inconsistency in 3D reconstruction. Various approaches have been proposed to address these challenges. LocalRF \cite{localrf} adopts progressive optimization, VastGaussian \cite{vastgaussian} partitions large scenes, Scaffold-GS \cite{scaffoldgs} introduces anchor-based Gaussian representations, and LongSplat \cite{longsplat} achieves robust unposed video reconstruction through incremental joint optimization. Traditional methods like COLMAP \cite{colmap} and SLAM systems \cite{orbslam, droidslam} explicitly estimate poses, while neural rendering methods such as BARF \cite{barf}, NoPe-NeRF \cite{nopenerf}, and CF-3DGS \cite{cf3dgs} jointly learn geometry and poses. However, existing methods largely overlook another critical characteristic of real-world videos: lossy compression. Most videos undergo compression during smartphone storage or online uploads, yet the impact of compression artifacts on 3D reconstruction remains underexplored.

\noindent\textbf{Video Compression in 3D Vision Tasks.}
Most videos captured or distributed in the real world are compressed in a lossy manner due to storage and transmission constraints, inevitably introducing artifacts. Furthermore, encoder design including different frame types and QP variation among frames induce per-frame quality fluctuations across a sequence. Consequently, in 2D vision, a large body of compression-aware restoration methods addresses these issues by explicitly accounting for compression artifacts or exploiting multi-frame cues to counter quality oscillations \cite{ alvarez2017compression, yang2018multi, feng2019coding, wang2019fast, li2021comisr, wang2023compression, zhai2023object, zhang2023video, he2025multi, yue2025survey, zeng2025plug}. In particular, many studies recently utilize coding priors to adaptively control the gain of each channel and to adjust restoration weights according to compression characteristics \cite{zhang2022dcngan, yang2023pimnet, zhu2024cpga, dehaghi2025reversing, liu2025coding, zhang2025qp}. In 3D vision, however, very few studies have examined the impact of video compression on reconstruction quality or explicitly modeled frame-level quality variations caused by codec design. This gap is particularly evident in NVS. In the NeRF domain, recent efforts have explored robustness to degraded or noisy inputs \cite{wang2022nerf, pearl2022nan, huang2023refsr, zhou2023nerflix, zhou2023nerflix2}, but these typically focus on improving overall input image quality rather than addressing compression-induced artifacts or frame-level quality variations that directly affect the 3D reconstruction process. In the context of 3DGS, HQGS \cite{linhqgs} enhances 3DGS robustness to degradations such as blur and coding artifact, but it does not address temporal inconsistencies or frame-level quality fluctuations, which is inherently prone to appear in compressed videos. As a result, its applicability to long, compressed video inputs is limited.

\paragraph{Position of Our Work.}
Existing NVS and 3DGS-based reconstruction methods mainly assume clean inputs and do not explicitly account for codec-induced frame quality variations, while compression-aware approaches in 2D vision focus on image restoration rather than stabilizing long-sequence 3D optimization. Our work bridges these directions by introducing a compression-aware framework for long-video NVS that operates directly on compressed and unposed real-world videos and explicitly leverages codec information to model frame-wise reliability. In particular, our CompSplat complements prior unposed 3DGS methods such as LongSplat~\cite{longsplat} and CF-3DGS~\cite{cf3dgs} by integrating compression indicators and training stability cues into the optimization process, enabling robust geometry and pose reconstruction under practical real-world compression conditions. 

%% file: sec/3_preliminaries.tex
\section{Preliminaries}
\label{sec:Preliminaries}

\noindent\textbf{3D Gaussian Splatting.}
In 3D Gaussian Splatting~\cite{3dgs}, 
a scene is represented as a set of anisotropic 3D Gaussians,
each parameterized by a mean $\boldsymbol{\mu} \in \mathbb{R}^3$, 
a covariance matrix $\boldsymbol{\Sigma} = \mathbf{R} \mathbf{S} \mathbf{S}^\top \mathbf{R}^\top$, 
and appearance attributes such as color and opacity $\alpha$. 
During rendering, each Gaussian is projected onto the image plane through the camera pose $\mathbf{W}$, 
and the final pixel value is obtained by alpha blending along the ray:
\begin{equation}
\label{eq:1}
C = \sum_i c_i \alpha_i \prod_{j < i}(1 - \alpha_j).
\end{equation}

\noindent\textbf{Anchor-based Representation.}
Building on the anchor-based Scaffold-GS~\cite{scaffoldgs}, LongSplat~\cite{longsplat} associates each anchor $v$ at position $\mathbf{x}_v$ with $k$ Gaussians whose centers are defined by relative offsets $\{\mathbf{O}_i\}$ and a scale factor $l_v$:
\begin{equation}
\label{eq:2}
\boldsymbol{\mu}_i = \mathbf{x}_v + \mathbf{O}_i \cdot l_v.
\end{equation}
Gaussian attributes are predicted from anchor features using lightweight MLP heads.


\noindent\textbf{Optimization Pipeline.}
LongSplat~\cite{longsplat} reconstructs unposed long videos through an incremental optimization pipeline composed of three stages: (a) \textit{Pose Estimation}, which aligns each frame via correspondence-guided PnP and photometric refinement; (b) \textit{Local Optimization}, which updates visible Gaussians within a short temporal window to preserve local consistency; and (c) \textit{Global Refinement}, which periodically optimizes all Gaussians and poses for sequence-wide coherence. The overall objective combines photometric, depth, and reprojection losses: 

\begin{equation}
\label{eq:3}
L_{\text{total}} = L_{\text{photo}} + L_{\text{depth}} + L_{\text{reprojection}}.
\end{equation}
Gaussian densification and pruning are further applied to regulate primitive density throughout training.

%% file: sec/4_method.tex
\section{Method}
\label{sec:Method}
We propose a compression-aware pipeline for long-video novel view synthesis that adapts the 3D Gaussian optimization process to frame-wise quality variations inherent in compressed videos.
As shown in Fig.~\ref{fig:2}, the pipeline consists of two components: (i) \textit{Quality-guided Density Control}, which modulates Gaussian densification and pruning based on codec-derived frame reliability,  and (ii) \textit{Quality Gap-aware Masking}, which mitigates view-level quality disparities by applying keypoint-driven pixel masking to the photometric supervision.
Together, these components mitigate quality imbalance across frames and yield stable, high-fidelity reconstruction under practical compression settings. 

\begin{figure}[t]
    \centering
    \includegraphics[width=\linewidth]{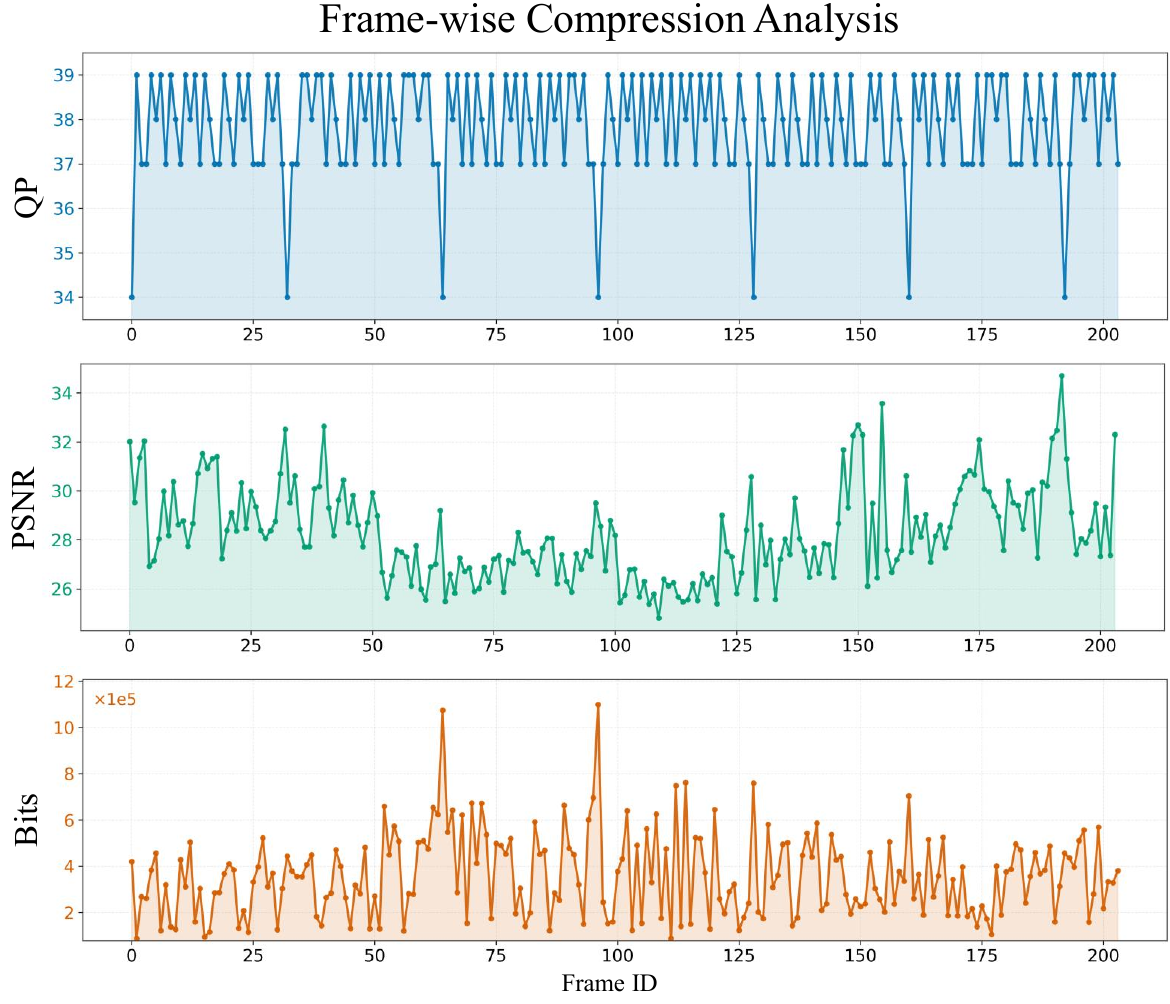}
    \caption{\textbf{Frame-wise compression analysis.} During video compression, each frame is encoded with QP values, leading to substantial inter-frame variations in PSNR and bitrate. This non-uniformity becomes more pronounced in long real-world videos, highlighting the need to explicitly consider frame-wise compression artifacts when applying 3DGS to compressed video.}
    \label{fig:3}
\end{figure}

\subsection{Problem Definition and Objectives}
\label{sec:problem}
\noindent\textbf{Problem Definition.}
Video compression removes inter-frame redundancy under bitrate constraints through intra- and inter-prediction. As illustrated in Fig.~\ref{fig:3}, these codec-driven variations induce substantial fluctuations in bit allocation and PSNR across a sequence, leading to inconsistent preservation of high-frequency details. Such frame-wise quality inconsistency leads to two major issues in existing unposed-3DGS frameworks.
\textit{First}, density optimization becomes unreliable: fixed densification and pruning thresholds treat all frames equally, causing high-quality frames to be over-pruned and low-quality frames to introduce noisy Gaussians.
\textit{Second}, quality gaps between adjacent frames degrade feature matching and pose estimation reliability, yielding unstable view supervision and inconsistent updates during incremental optimization.

\noindent\textbf{Objectives.}
To address these two issues, we develop a compression-aware pipeline that adapts Gaussian optimization process to frame-level compression characteristics:
(i) We regulate Gaussian densification and pruning using a frame confidence score, enabling density control that responds to quality variations across frames; and
(ii) We mitigate quality gaps by modulating photometric supervision with a view-level gap score, reducing the influence of unreliable frames.
Together, these objectives ensure consistent density evolution and stable view supervision in compressed long-video novel view synthesis. 

\subsection{Quality-guided Density Control}
\noindent\textbf{Adaptive Frame Quality Estimation.}
Mordern video codecs can assign different bits and QP to each frame at the cost of quality fluctuation per frame. Obviously, frames with lower QP values retain more high-frequency details, whereas frames with higher QP values tend to lose fine structures. Similarly, frames with higher bitrates means that more information is needed due to increased scene complexity. To quantify these characteristics, we newly introduce a QP-based confidence score $q_t^q$ and a bitrate-based confidence score $q_t^b$, and then we combine them into a unified frame confidence score $q_t = q_t^q + q_t^b$. Both confidence terms are computed through linear normalization of the respective value over the sequence:
\begin{equation}
\label{eq:4}
q_t^{q}
= \lambda^q \frac{Q_{\max}^{f} - Q_{t}^{f}}{Q_{\max}^{f} - Q_{\min}^{f} + \varepsilon},
\quad
q_t^{b}
= \lambda^b \frac{B_{t}^{f} - B_{\min}^{f}}{B_{\max}^{f} - B_{\min}^{f} + \varepsilon}.
\end{equation}
where $Q_t^{f}$ and $B_t^{f}$ denote the QP and the bits at the corresponding frame $t$, respectively. $\varepsilon$ is a small constant to avoid division by zero, and $\lambda^{q}$ and $\lambda^{b}$ are weighting hyperparameters controlling the QP and bitrate contributions, of frame $t$, respectively. $Q_{\min}^{f}$, $Q_{\max}^{f}$, $B_{\min}^{f}$, and $B_{\max}^{f}$ represent the minimum and maximum values across the sequence. To ensure temporal stability across frames, an exponential moving average (EMA) is applied:
\begin{equation}
\label{eq:5}
\bar{q}_t = \beta \bar{q}_{t-1} + (1 - \beta) q_t,
\end{equation}
where $\beta$ denotes the momentum parameter. The resulting EMA confidence $\bar{q}_t$ serves as a baseline for scale-based pruning and adaptive Gaussian density control in subsequent stages.

\noindent\textbf{Gaussian Scale-based Pruning.}
In GS optimization, the spatial density of primitives is determined by two criteria: 
(i) new Gaussians are densified when their accumulated gradient exceeds a threshold $\theta$, 
and (ii) existing ones are pruned when their opacity $\alpha$ falls below a threshold $\omega$:

\begin{equation}
\label{eq:6}
\text{densify if } g_t > \theta \quad \text{and} \quad \text{prune if } \alpha_t < \omega,
\end{equation}
where $g_t$ and $\alpha_t$ denote the gradient magnitude and opacity of a Gaussian at frame~$t$, respectively.  
In our pipline, scale-based pruning modifies only the pruning rule by making $\omega$ responsive to Gaussian scale.

Compressed frames often lose high-frequency components such as fine textures and edges, producing blurred regions that generate over-diffused Gaussians. To filter out these unreliable primitives, we introduce a scale-based pruning step. For each anchor $v$, we compute a representative scale by averaging its Gaussian offsets and normalizing it by the median scale in the scene:
\begin{equation}
\label{eq:7}
U_v = \|\mathbf{s}^{(v)}\|_2, \qquad 
\tilde{U}_v = \frac{U_v}{\mathrm{median}(U)}.
\end{equation}
Large normalized scales $\tilde{U}_v$ indicate that the Gaussians are likely originating from degraded frames. To couple scale filtering with frame reliability, we adaptively modulate the pruning threshold \textit{per frame} as:
\begin{equation}
\label{eq:8}
\omega_t = \omega_0 \exp(\bar{q}_t \tilde{U}_v),
\end{equation}
where $\omega_0$ is the base opacity threshold and $\bar{q}_t$ is the EMA confidence from Eq.~\ref{eq:5}.  
This formulation enables frame-adaptive removal of over-scaled Gaussians while preserving compact, reliable primitives.

\noindent\textbf{Adaptive Densification and Pruning.}
Once over-diffused Gaussians are removed through scale-based pruning, reliable primitives receive clearer gradient signals, allowing density evolution to respond more faithfully to frame-wise quality variations. We therefore adaptively modulate the thresholds $\theta$ and $\omega$ using frame confidence to stabilize densification and pruning under varying compression levels.  
Given the current confidence $q_t$ and its temporal average $\bar{q}_t$, we define:
\begin{equation}
\label{eq:9}
\theta_t = \theta_0 \exp(\bar{q}_t - q_t), \qquad
\omega'_t = \omega_0 \exp(\bar{q}_t - q_t),
\end{equation}
where $\theta_0$ and $\omega_0$ are base thresholds.
When $q_t > \bar{q}_t$, the model reduces $\theta_t$ to allow denser Gaussian creation and decreases $\omega'_t$ to relax pruning.  
Conversely, $q_t < \bar{q}_t$ increases both thresholds, suppressing Gaussian creation and enforcing stronger pruning. This confidence-driven modulation of $\theta$ and $\omega$ stabilizes density evolution and prevents inconsistent Gaussian growth across frames with varying compression quality.

\begin{table*}[t]
\centering
\small
\setlength{\tabcolsep}{3.2pt}
\renewcommand{\arraystretch}{1.15}
\caption{\textbf{Quantitative comparison on the Free dataset~\cite{free} across various baseline methods.} “OOM” indicates an out-of-memory failure. “Uncomp.” refers to training and evaluation on original uncompressed videos. Our method consistently achieves higher PSNR, SSIM, and LPIPS across all scenes by explicitly accounting for compressed video artifacts, leading to more stable and accurate reconstructions under challenging camera trajectories and complex geometric structures.
}
\label{tab:1}
\resizebox{\textwidth}{!}{%
\begin{tabular}{l!{\vrule width 0.6pt}ccc|ccc|ccc|ccc|ccc|ccc|ccc}
\toprule
\multirow{2}{*}{Scenes} &
\multicolumn{3}{c|}{\textcolor{gray}{LongSplat (Uncomp.)}} &
\multicolumn{3}{c|}{NoPe-NeRF~\cite{nopenerf}} &
\multicolumn{3}{c|}{LocalRF~\cite{localrf}} &
\multicolumn{3}{c|}{CF-3DGS~\cite{cf3dgs}} &
\multicolumn{3}{c|}{CF-3DGS + \textbf{Ours}} &
\multicolumn{3}{c|}{LongSplat~\cite{longsplat}} &
\multicolumn{3}{c}{LongSplat + \textbf{Ours}} \\
\cmidrule(lr){2-4} \cmidrule(lr){5-7} \cmidrule(lr){8-10} \cmidrule(lr){11-13} \cmidrule(lr){14-16} \cmidrule(lr){17-19} \cmidrule(lr){20-22}
& PSNR$\uparrow$ & SSIM$\uparrow$ & LPIPS$\downarrow$ & 
PSNR$\uparrow$ & SSIM$\uparrow$ & LPIPS$\downarrow$ & 
PSNR$\uparrow$ & SSIM$\uparrow$ & LPIPS$\downarrow$ & 
PSNR$\uparrow$ & SSIM$\uparrow$ & LPIPS$\downarrow$ & 
PSNR$\uparrow$ & SSIM$\uparrow$ & LPIPS$\downarrow$ & 
PSNR$\uparrow$ & SSIM$\uparrow$ & LPIPS$\downarrow$ & 
PSNR$\uparrow$ & SSIM$\uparrow$ & LPIPS$\downarrow$ \\
\midrule
Grass   & \textcolor{gray}{26.03} & \textcolor{gray}{0.79} & \textcolor{gray}{0.21} & 17.84 & 0.41 & 0.73 & 20.20 & 0.47 & 0.44 & OOM   & OOM  & OOM  & OOM  & OOM  & OOM  & 23.61 & 0.64 & 0.33 & \textbf{24.13} & \textbf{0.66} & \textbf{0.30} \\
Hydrant & \textcolor{gray}{23.23} & \textcolor{gray}{0.73} & \textcolor{gray}{0.21} & 19.43 & 0.55 & 0.59 & 20.34 & 0.59 & 0.31 & 12.63 & 0.33 & 0.61 & 9.19 & 0.19 & 0.66 & 23.55 & 0.72 & 0.27 & \textbf{24.28} & \textbf{0.67} & \textbf{0.30} \\
Lab     & \textcolor{gray}{26.36} & \textcolor{gray}{0.86} & \textcolor{gray}{0.16} & 17.61 & 0.55 & 0.61 & 17.25 & 0.55 & 0.31 & OOM   & OOM  & OOM  & OOM  & OOM  & OOM  & 26.38 & \textbf{0.86} & 0.18 & \textbf{26.87} & \textbf{0.86} & \textbf{0.17} \\
Pillar  & \textcolor{gray}{28.83} & \textcolor{gray}{0.84} & \textcolor{gray}{0.19} & 19.07 & 0.59 & 0.62 & 25.74 & 0.70 & 0.34 & 13.92 & 0.41 & 0.66 & 14.78 & 0.39 & 0.64 & 26.15 & 0.71 & 0.34 & \textbf{26.96} & \textbf{0.74} & \textbf{0.32} \\
Road    & \textcolor{gray}{21.54} & \textcolor{gray}{0.60} & \textcolor{gray}{0.34} & 19.73 & 0.63 & 0.62 & 22.54 & 0.64 & 0.38 & OOM   & OOM  & OOM  & OOM & OOM & OOM & 20.91 & 0.59 & 0.43 & \textbf{21.75} & \textbf{0.62} & \textbf{0.41} \\
Sky     & \textcolor{gray}{26.67} & \textcolor{gray}{0.86} & \textcolor{gray}{0.19} & 16.08 & 0.59 & 0.58 & 19.56 & 0.65 & 0.28 & OOM & OOM & OOM & OOM  & OOM  & OOM  & 25.28 & 0.82 & 0.27 & \textbf{25.77} & \textbf{0.83} & \textbf{0.26} \\
Stair   & \textcolor{gray}{29.21} & \textcolor{gray}{0.84} & \textcolor{gray}{0.19} & 20.55 & 0.62 & 0.59 & 26.08 & 0.81 & 0.20 & 14.32 & 0.43 & 0.60 & 12.01 & 0.34 & 0.63 & 26.97 & 0.77 & 0.29 & \textbf{27.41} & \textbf{0.78} & \textbf{0.28} \\
\midrule
Avg     & \textcolor{gray}{25.98} & \textcolor{gray}{0.79} & \textcolor{gray}{0.21} & 18.62 & 0.56 & 0.62 & 21.96 & 0.63 & 0.32 & 13.62    & 0.39   & 0.62   & 12.00   & 0.31   & 0.65   & 24.69 & 0.73 & 0.30 & \textbf{25.31} & \textbf{0.74} & \textbf{0.29} \\
\bottomrule
\end{tabular}%
}
\end{table*}

\subsection{Quality Gap-aware Masking}
Frame-level compression variability creates quality gaps (\cref{fig:3}) that adversely affect view supervision, corresponding to the second objective described in Sec.~\ref{sec:problem}. In LongSplat~\cite{longsplat}, each incoming frame is aligned through MASt3R\cite{mast3r}-based keypoint matching followed by PnP initialization, after which newly observed regions are unprojected to expand the Gaussian set. However, when frames suffer from compression artifacts or reduced detail, the number of detected keypoints decreases, resulting in a less number of geometrically consistent correspondences. This degradation reduces pose reliability and destabilizes the subsequent visibility-adapted local optimization.

To address this quality gap, we compute an inlier ratio for each view $t$ using the number of detected keypoints $K_t$ and the number of inlier correspondences $I_t$ returned by MASt3R~\cite{mast3r}:
\begin{equation}
\label{eq:10}
r_t = \frac{I_t}{K_t + \varepsilon},
\end{equation}
where $\varepsilon$ is a small constant to prevent division by zero.
This ratio serves as a view-level reliability measure: high-quality frames generally produce more consistent matches, whereas degraded frames yield low inlier ratios due to loss of structure from compression.

To reduce the influence of these unreliable views, we convert the inlier ratio into a pixel drop rate:
\begin{equation}
\label{eq:11}
d_t = \eta (1 - r_t),
\end{equation}
where $\eta$ controls the sensitivity to quality gaps. Then, a Bernoulli mask with probability $d_t$ is applied to the photometric loss, discarding a proportion of pixels in accordance with frame reliability. Consequently, reliable views provide stronger supervision, while less reliable views contribute less, yielding stable view-level optimization despite large frame-to-frame quality differences.

%% file: sec/5_experiments.tex
\section{Experiments}

\subsection{Experimental Setup}

\noindent\textbf{Compression Settings.}
 To reflect real-world encoding environments as in~\cite{antsiferova2022video}, all videos were encoded using x265 codec \cite{x265} at 60~fps with YUV~4:2:0 color sampling. A GOP length of 32 and a Random Access coding structure were adopted, while Open-GOP was disabled. We used a compression level corresponding to QP 37 as common practice to evaluate the quality \cite{he2025rivuletmlp}, and more QP-variants can be found at supplementary material.

\begin{figure*}[t]
    \centering
    \includegraphics[width=\textwidth]{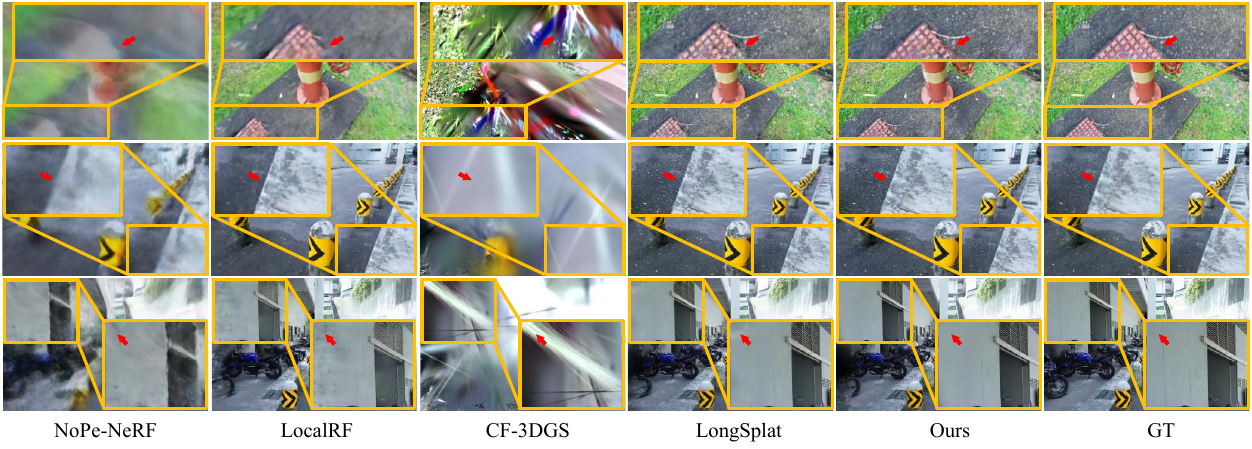}
    \vspace{-0.7cm}
    \caption{\textbf{Qualitative comparison on the Free dataset~\cite{free}}. We compare our method against CF-3DGS \cite{cf3dgs}, NoPe-NeRF \cite{nopenerf}, LocalRF \cite{localrf}, and LongSplat \cite{longsplat}. CF-3DGS produces highly diffused or incorrect Gaussians on compressed datasets, making object and scene structures difficult to recognize. Other baseline methods also yield blurry or geometrically distorted reconstructions. LongSplat performs relatively well; however, when combined with our approach, the results exhibit sharper textures and clearer object boundaries, demonstrating improved reconstruction quality even under compressed conditions.}
    \label{fig:4}
\end{figure*}

\begin{table}[t]
\centering
\footnotesize
\setlength{\tabcolsep}{6pt}
\renewcommand{\arraystretch}{1.1}
\caption{\textbf{Quantitative evaluation of camera pose estimation accuracy on the Free dataset~\cite{free}.} 
}
\label{tab:2}
\begin{tabular}{l!{\vrule width 0.6pt}ccc}
\toprule
Method & RPE$_t\downarrow$ & RPE$_r\downarrow$ & ATE$\downarrow$ \\
\midrule
\textcolor{gray}{LongSplat (Uncomp.)} & \textcolor{gray}{0.616} & \textcolor{gray}{1.447} & \textcolor{gray}{0.010} \\
\midrule
NoPe-NeRF              & 5.834   & 4.716   & 0.545   \\
CF-3DGS                & 0.807   & 6.241   & 0.030   \\
CF-3DGS + \textbf{Ours}         & 0.824   & 6.709   & 0.029   \\
LongSplat              & 0.872   & 1.721   & 0.016   \\
LongSplat + \textbf{Ours}       & \textbf{0.539} & \textbf{1.047} & \textbf{0.008} \\
\bottomrule
\end{tabular}
\vspace{-1pt}
\end{table}

\begin{table}[t]
\centering
\small
\setlength{\tabcolsep}{3pt}
\renewcommand{\arraystretch}{1.12}
\caption{\textbf{Quantitative evaluation of novel view synthesis quality on the Tanks and Temples dataset~\cite{tankandtemples}.}
\emph{Bold indicates the best \underline{within each baseline +Ours pair}}}
\label{tab:3}
\resizebox{\columnwidth}{!}{%
\begin{tabular}{l!{\vrule width 0.6pt}cccccc}
\toprule
Method & PSNR$\uparrow$ & SSIM$\uparrow$ & LPIPS$\downarrow$ & RPE$_t\downarrow$ & RPE$_r\downarrow$ & ATE$\downarrow$ \\
\midrule
CF-3DGS                 & 27.36 & 0.80 & \textbf{0.27} & 0.052 & 0.099 & 0.006 \\
CF-3DGS + \textbf{Ours} & \textbf{27.84} & \textbf{0.82} & \textbf{0.27} & \textbf{0.049} & \textbf{0.085} & \textbf{0.005} \\
\midrule
LongSplat               & 28.21 & \textbf{0.82} & \textbf{0.26} & 2.265 & 0.831 & 0.031 \\
LongSplat + \textbf{Ours} & \textbf{28.37} & \textbf{0.82} & \textbf{0.26} & \textbf{2.176} & \textbf{0.823} & \textbf{0.029} \\
\bottomrule
\end{tabular}%
}
\end{table}

\begin{table}[t]
\centering
\small
\setlength{\tabcolsep}{3pt}
\renewcommand{\arraystretch}{1.12}
\caption{\textbf{Quantitative evaluation on the Hike dataset.}
Comparison between LongSplat and our method in terms of photometric quality and pose accuracy.}
\label{tab:4}
\resizebox{0.85\columnwidth}{!}{%
\begin{tabular}{l!{\vrule width 0.6pt}cccccc}
\toprule
Method
& PSNR$\uparrow$
& SSIM$\uparrow$
& LPIPS$\downarrow$
& RPE$_t\downarrow$
& RPE$_r\downarrow$
& ATE$\downarrow$ \\
\midrule
LongSplat
& 19.37
& 0.52
& 0.36
& 1.180
& 6.639
& 0.025 \\
Ours
& \textbf{19.65}
& \textbf{0.54}
& \textbf{0.35}
& \textbf{1.152}
& \textbf{6.185}
& \textbf{0.024} \\
\bottomrule
\end{tabular}%
}
\end{table}

\noindent\textbf{Datasets.}
Using the compression configuration described, we evaluate our pipeline on three real-world video datasets. 

• \textbf{Tanks and Temples}~\cite{tankandtemples}: Eight forward-facing scenes; trained at full resolution and evaluated on every 9th frame.

• \textbf{Free Dataset}~\cite{free}: Seven handheld videos with unconstrained camera motion; trained at half resolution and evaluated on every 9th frame.

• \textbf{Hike Dataset}~\cite{hike}: Long outdoor videos with complex motion and large-scale geometry; trained at 1/4 resolution with half the frames, evaluated every 10th frame.

\noindent\textbf{Baselines.}
We compare our method with unposed reconstruction approaches, including NoPe-NeRF \cite{nopenerf}, LocalRF \cite{localrf}, CF-3DGS \cite{cf3dgs}, and LongSplat \cite{longsplat}.

\noindent\textbf{Implementation Details.}
We adopt LongSplat \cite{longsplat} and CF-3DGS \cite{cf3dgs} as our baseline framework and integrate our proposed method, CompSplat, within the pipeline. Following the original LongSplat configuration, the number of iterations for the local, global, and refine stages is set to 400, 900, and 10,000, respectively. All experiments are conducted on an NVIDIA 3080 GPU, while CF-3DGS is trained on an NVIDIA A6000 GPU. For our method, the confidence weighting parameters are set to $\lambda^{q}=1.0$ and $\lambda^{b}=0.5$, the drop-rate scaling parameter to $\eta=0.5$, and the numerical stability constant to $\varepsilon=10^{-6}$.

\subsection{Comparisons}
\begin{figure}[t]
\centering
\includegraphics[width=\columnwidth]{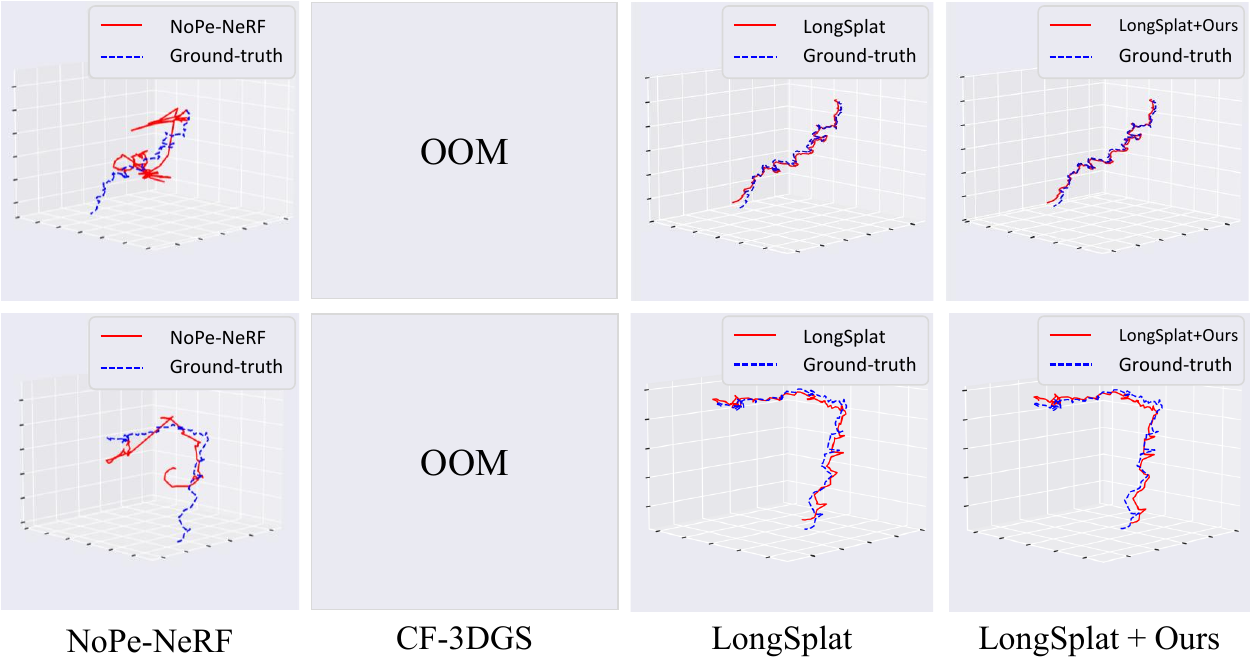}
\caption{\textbf{Visualization of camera trajectories on the Free dataset \cite{free}.} CF-3DGS fails to estimate a valid camera trajectory due to OOM when processing long sequences. Compared to LongSplat and other baselines, our method produces more stable and consistent camera trajectories under compressed video frames.}

\label{fig:5}
\end{figure}

\noindent\textbf{Free Dataset.}
\cref{tab:1,tab:2} show that our method achieves state-of-the-art performance in both rendering quality and camera pose estimation. \cref{fig:4,fig:5} demonstrate that our method better preserves fine textures and object boundaries and provides more accurate camera trajectory estimation, resulting in clearer and more stable reconstructions than existing baselines.



\begin{figure}[t]
\centering
\includegraphics[width=\columnwidth]{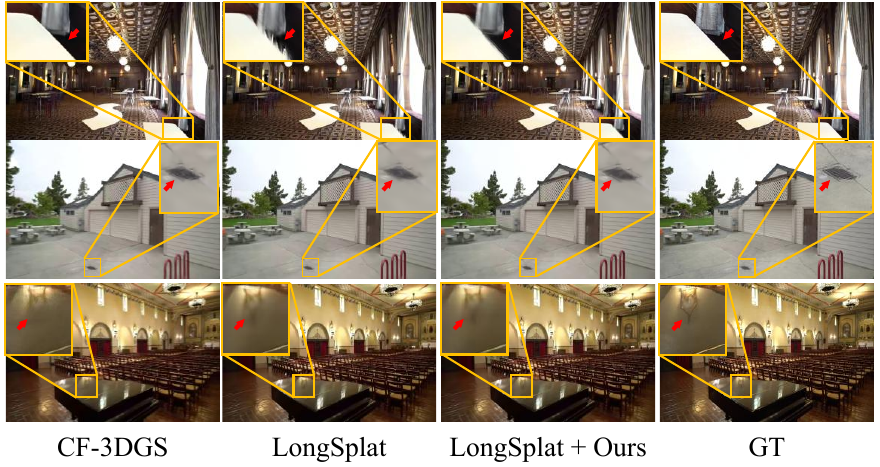}
\caption{\textbf{Qualitative results on Tank and Temples dataset \cite{tankandtemples}.} Comparison of rendered views across various method with ours.}
\label{fig:6}
\end{figure}

\noindent\textbf{Tank and Temples Dataset.}
\cref{tab:3} presents the quantitative results on the Tanks and Temples dataset~\cite{tankandtemples}. Across all evaluated baselines, adding our method consistently improves the results, indicating that our approach generalizes well and provides stable gains on this benchmark. \cref{fig:6} further shows that our method preserves fine geometric details and maintains texture consistency in challenging regions such as reflective surfaces and thin structural elements.

\noindent\textbf{Hike Dataset.}
Tab.~\ref{tab:4} compares LongSplat~\cite{longsplat} and our method on the Hike dataset~\cite{hike}. Our method consistently outperforms the LongSplat baseline across all photometric and pose metrics, demonstrating robust performance even in complex scenes. Please refer to the Supplementary Material for the full per-scene quantitative results.

\subsection{Ablation Study}

\noindent\textbf{Ablation on optimization components.}
\cref{tab:5} reports an ablation study of three components: (a) Gaussian Scale-based Pruning, (b) Adaptive Densification and Pruning, and (c) Quality Gap-aware Masking. When each component is applied individually, the performance improves to varying degrees in both photometric and camera metrics. Combining two components yields comparable or slightly improved results, while using all components (Ours) achieves the best performance across all metrics.

\begin{table}[t]
\centering
\small
\setlength{\tabcolsep}{6pt} 
\renewcommand{\arraystretch}{1.2}

\caption{\textbf{Ablation on Compression-aware optimization components.} Effects of (a) Gaussian Scale-based Pruning, (b) Adaptive Densification and Pruning, and (c) Quality Gap-aware Masking, averaged over all Free dataset~\cite{free} scenes.}


\label{tab:5}

\resizebox{\columnwidth}{!}{%
\begin{tabular}{ccc|cccccc}
\toprule
\multicolumn{3}{c|}{\textbf{Method}} & \multirow{2}{*}{\textbf{PSNR$\uparrow$}} & \multirow{2}{*}{\textbf{SSIM$\uparrow$}} & \multirow{2}{*}{\textbf{LPIPS$\downarrow$}} & \multirow{2}{*}{\textbf{RPE$_t$$\downarrow$}} & \multirow{2}{*}{\textbf{RPE$_r$$\downarrow$}} & \multirow{2}{*}{\textbf{ATE$\downarrow$}} \\
\cmidrule(lr){1-3}
(a) & (b) & (c) & \multicolumn{6}{c}{} \\ 
\midrule
 &  &  & 24.69 & 0.73 & 0.30 & 0.872 & 1.721 & 0.016 \\   
\checkmark &  &  & 24.96 & \textbf{0.74} & \textbf{0.29} & 0.680 & 1.566 & 0.012 \\
& \checkmark &  & 24.91 & 0.73 & \textbf{0.29} & 0.716 & 1.171 & 0.012 \\
 &  & \checkmark & 24.92 & \textbf{0.74} & 0.30 & 0.782 & 1.513 & 0.011 \\
\midrule
\checkmark & \checkmark &  & 24.83 & 0.73 & 0.30 & 0.759 & 1.580 & 0.013 \\
\checkmark &  & \checkmark & 24.73 & 0.73 & 0.30 & 0.708 & 1.707 & 0.012 \\
 & \checkmark & \checkmark & 24.72 & 0.73 & 0.30 & 0.802 & 1.405 & 0.012 \\
\midrule
\checkmark & \checkmark & \checkmark & \textbf{25.31} & \textbf{0.74} & \textbf{0.29} & \textbf{0.539} & \textbf{1.047} & \textbf{0.008} \\
\bottomrule
\end{tabular}
}

\end{table}

\begin{table}[t]
\centering
\footnotesize
\setlength{\tabcolsep}{3pt}
\renewcommand{\arraystretch}{1.2}
\caption{\textbf{Ablation of frame-quality estimation components on the Free dataset~\cite{free}.} 
QP-only and bitrate-only scoring both reduce accuracy, while the combined formulation yields the best photometric and pose results.}
\label{tab:6}
\resizebox{\columnwidth}{!}{%
\begin{tabular}{l!{\vrule width 0.6pt}cccccc}
\toprule
Method   & PSNR$\uparrow$ & SSIM$\uparrow$ & LPIPS$\downarrow$ & RPE$_t\downarrow$ & RPE$_r\downarrow$ & ATE$\downarrow$ \\
\midrule
Bit-only & 24.81 & 0.73 & \textbf{0.29} & 0.882 & 2.086 & 0.015 \\
QP-only  & 21.92 & 0.62 & 0.36 & 1.179 & 4.111 & 0.017 \\
\midrule
Together (Ours)  & \textbf{25.31} & \textbf{0.74} & \textbf{0.29} & \textbf{0.539} & \textbf{1.047} & \textbf{0.008} \\
\bottomrule
\end{tabular}%
}
\end{table}

\noindent\textbf{Ablation on Frame-Quality Estimation.}
\cref{tab:6} analyzes frame-quality confidence based on QP only or bitrate only. Both single-factor designs degrade performance on photometric and camera metrics, with QP-only being particularly limited due to its coarse, discrete nature. In contrast, combining QP and bitrate yields the best overall results.


\noindent\textbf{Video Codec Comparison.}
Tab.~\ref{tab:7} evaluates performance across widely used video codecs from different generations, including VVC, H.264, and H.265. Our method consistently outperforms LongSplat across all codecs, indicating stable performance under different compression standards. 

\begin{figure}[t]
\centering
\includegraphics[width=\columnwidth]{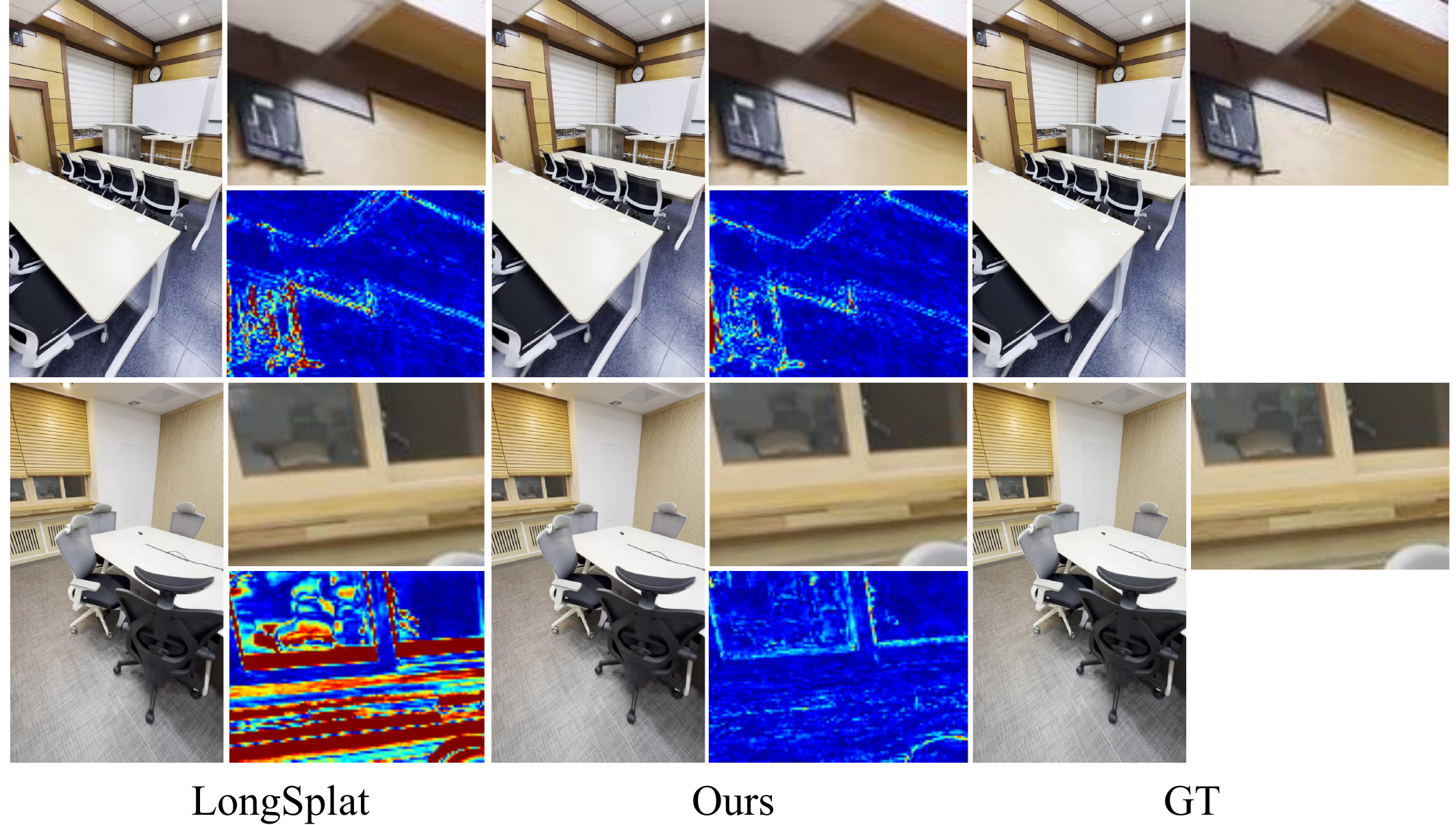}

\caption{\textbf{Qualitative results on real-world compressed videos.} Comparison of rendered views between LongSplat and ours. Heatmaps show per-pixel deviations from GT, with warmer colors indicating larger differences.}

\label{fig:7}
\end{figure}

\noindent\textbf{Real-World Compressed Video Evaluation.}
\cref{fig:7,tab:8} present qualitative and quantitative comparisons on real-world videos captured using a mobile device and compressed under our setting. Our method outperforms LongSplat across all metrics. Qualitatively, our results better preserve textures and edge details under compression, as evidenced by the error heatmaps in \cref{fig:7}, where the rendered results of our method exhibit fewer warm-colored regions, indicating smaller deviations from the ground truth.
%
%

\captionsetup{font=small, labelfont=bf}
\newcommand{\six}{-- & -- & -- & -- & -- & --}

\begin{table}[t]
\centering
\small
\setlength{\tabcolsep}{3pt}
\renewcommand{\arraystretch}{1.0}
\caption{\textbf{Comparison across different video codecs.}
Evaluation on the Free dataset~\cite{free} under VVC, H.264, and H.265 (QP37) compression, comparing LongSplat and our method.}
\label{tab:7}

\resizebox{\columnwidth}{!}{%
\begin{tabular}{l|c|cccccc}
\toprule
\textbf{Method} & \textbf{Codec} &
PSNR$\uparrow$ & SSIM$\uparrow$ & LPIPS$\downarrow$ &
RPE$_t\downarrow$ & RPE$_r\downarrow$ & ATE$\downarrow$ \\
\midrule
LongSplat & \multirow{2}{*}{VVC}
& 25.21 & \textbf{0.73} & \textbf{0.33} & 0.595 & 1.419 & \textbf{0.010} \\
\textbf{Ours} &
& \textbf{25.35} & \textbf{0.73} & \textbf{0.33}
& \textbf{0.591} & \textbf{1.373} & 0.011 \\
\midrule
LongSplat & \multirow{2}{*}{H.264}
& 24.80 & 0.71 & \textbf{0.34} & 0.971 & 1.818 & 0.016 \\
\textbf{Ours} &
& \textbf{24.95} & \textbf{0.72} & \textbf{0.34}
& \textbf{0.856} & \textbf{1.589} & \textbf{0.014} \\
\midrule
LongSplat & \multirow{2}{*}{H.265}
& 24.69 & 0.73 & 0.30 & 0.872 & 1.721 & 0.016 \\
\textbf{Ours} &
& \textbf{25.31} & \textbf{0.74} & \textbf{0.29}
& \textbf{0.539} & \textbf{1.047} & \textbf{0.008} \\
\bottomrule
\end{tabular}%
}
\end{table}

\begin{table}[t]
\centering

\footnotesize
\setlength{\tabcolsep}{1.6pt}
\renewcommand{\arraystretch}{1.0}

\caption{\textbf{Quantitative comparison on real-world compressed videos.}
All sequences are captured using a mobile device and evaluated under the same compression settings.}
\label{tab:8}
\resizebox{0.85\columnwidth}{!}{%
\begin{tabular}{l l ccc ccc}
\toprule
Scenes & Method 
& PSNR$\uparrow$ & SSIM$\uparrow$ & LPIPS$\downarrow$
& RPE$_t\downarrow$ & RPE$_r\downarrow$ & ATE$\downarrow$ \\
\midrule

\multirow{2}{*}{Scene1}
& LongSplat
& 35.60 & 0.96 & 0.08
& 0.062 & 0.086 & 0.001 \\
& Ours
& \textbf{35.68} & \textbf{0.96} & \textbf{0.08}
& \textbf{0.061} & \textbf{0.077} & \textbf{0.001} \\
\midrule

\multirow{2}{*}{Scene2}
& LongSplat
& 31.43 & 0.86 & 0.09
& 0.135 & 0.190 & \textbf{0.002} \\
& Ours
& \textbf{31.47} & \textbf{0.86} & \textbf{0.09}
& \textbf{0.124} & \textbf{0.165} & \textbf{0.002} \\
\bottomrule
\end{tabular}%
}
\end{table}

%% file: sec/6_conclusion.tex
\section{Conclusion}
We presented CompSplat, a compression-aware 3DGS framework for real-world long and unposed video reconstruction. By explicitly modeling frame-wise compression characteristics, CompSplat adaptively regulates Gaussian densification and pruning, mitigating geometric drift and maintaining consistent reconstruction quality under varying compression levels. 
Extensive experiments demonstrate that CompSplat achieves state-of-the-art performance on real compressed videos, outperforming existing unposed 3DGS methods in both quality and stability. These results highlight the effectiveness of our approach in practical scenarios, advancing the robustness and generalization of 3DGS for real-world, practical bandwidth-constrained video applications. We believe this work lays an important foundation for compression-aware 3DGS in long unposed video reconstruction.

%% file: sec/X_suppl.tex
\clearpage
\setcounter{page}{1}

\setcounter{section}{0}

\setcounter{figure}{7}
\setcounter{table}{8}

\renewcommand{\thesection}{\Alph{section}}

\maketitlesupplementary

This supplementary material provides additional implementation details, extended analyses of compression-related behavior, further experimental results across datasets and settings, and additional qualitative visualizations to complement the main paper.

\section{Implementation Details}
\label{sec:implementation}



\paragraph{Real-world Dataset Setup.}
We collected a real-world video dataset by recording two indoor scenes using an iPhone 17 Pro Max at 4K resolution and 30~fps. To ensure consistent training resolution, each captured frame was downsampled by a factor of 4×, resulting in input images at 1/4 of the original 4K resolution. Each captured sequence was then compressed using \texttt{ffmpeg} with the \texttt{libx265} encoder, following the compression settings described above.

\paragraph{VVC Compression Setup.}
For the VVC experiments, we encoded each sequence using the official VVC 
reference software VTM~23.1 in the Random Access (RA) configuration. 
The RA setting was used with an intra period and GOP size of 32 frames, 
matching the temporal structure used in our HEVC/x265 experiments. 
We evaluated two compression levels using QP value of 37. 
The encoding was performed with the VTM \texttt{EncoderApp} and the standard 
\texttt{encoder\_randomaccess\_vtm.cfg} configuration file. 
After encoding, all bitstreams were decoded using \texttt{DecoderApp} to obtain 
the reconstructed YUV sequences, which were finally converted back into 
frame-wise RGB images using \texttt{ffmpeg} while preserving the original 
resolution and frame rate.

\begin{figure}[t]
\centering
\includegraphics[width=\columnwidth]{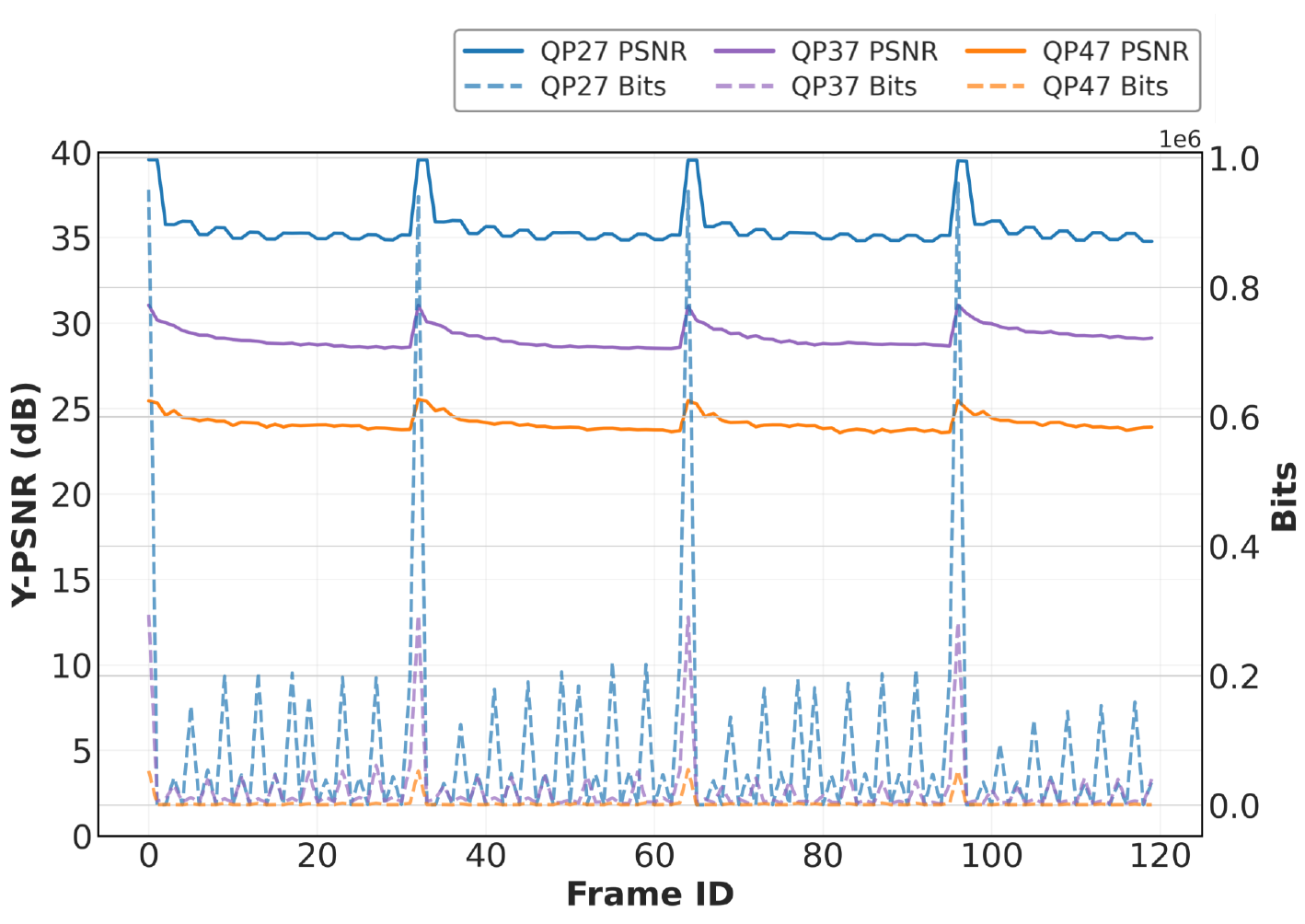}
\caption{\textbf{Frame-wise Y-PSNR and bits per frame of video sequences encoded at various QP values.} Lower QP values yield higher PSNR and larger bit usage, while higher QP values result in reduced quality and substantially fewer coded bits. Periodic peaks correspond to I-frames within the GOP structure.}
\label{fig:supp-fig8}
\end{figure}

\section{Analysis}
This section analyzes how compression-induced quality fluctuations manifest in real videos and provides empirical evidence for the issues identified in the main paper, thereby clarifying the motivation behind our compression-aware design. In this section, we focus on four aspects: 
(i) the impact of QP on input frames,
(ii) frame-quality estimation based on QP and bitrate,
(iii) the behavior of adaptive threshold adjustment for density control, and 
(iiii) the impact of PSNR gaps on feature matching stability.

\begin{figure*}[tbh]
\centering
\includegraphics[width=\textwidth]{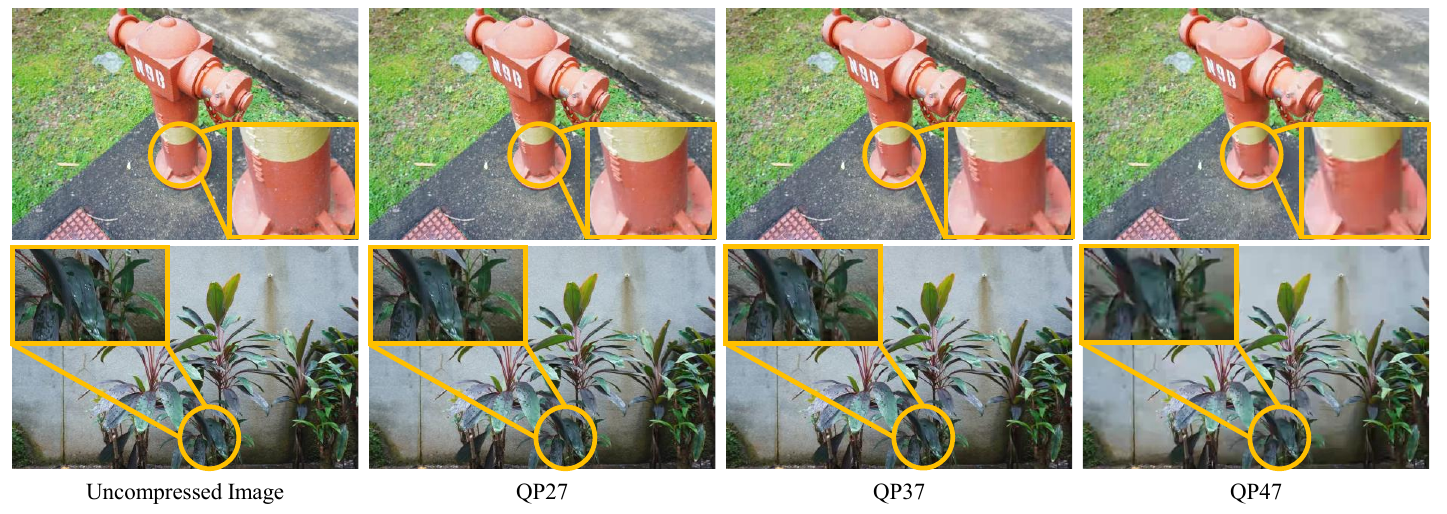}
\caption{\textbf{Effects of quantization parameter (QP) on video frames.} As QP increases from 27 to 47, overall frame quality progressively degrades compared to the uncompressed reference.}
\label{fig:supp-fig9}
\end{figure*}


\paragraph{Impact of QP on Input Frames.}
As shown in Fig.~\ref{fig:supp-fig8}, the per-frame PSNR and the bitrate vary significantly with the chosen QP. This indicates that not only the overall visual quality of the input frames, but also the degree of frame-to-frame quality variation within each sequence, depends strongly on the compression level. Fig.~\ref{fig:supp-fig9} illustrates the visual differences in the input frames at different QP values. As the QP increases from 27 to 47, the input frames become more heavily compressed, resulting in noticeable loss of fine details, increased blurring, and degradation in color fidelity.

\begin{figure}[t]
\centering
\includegraphics[width=\columnwidth]{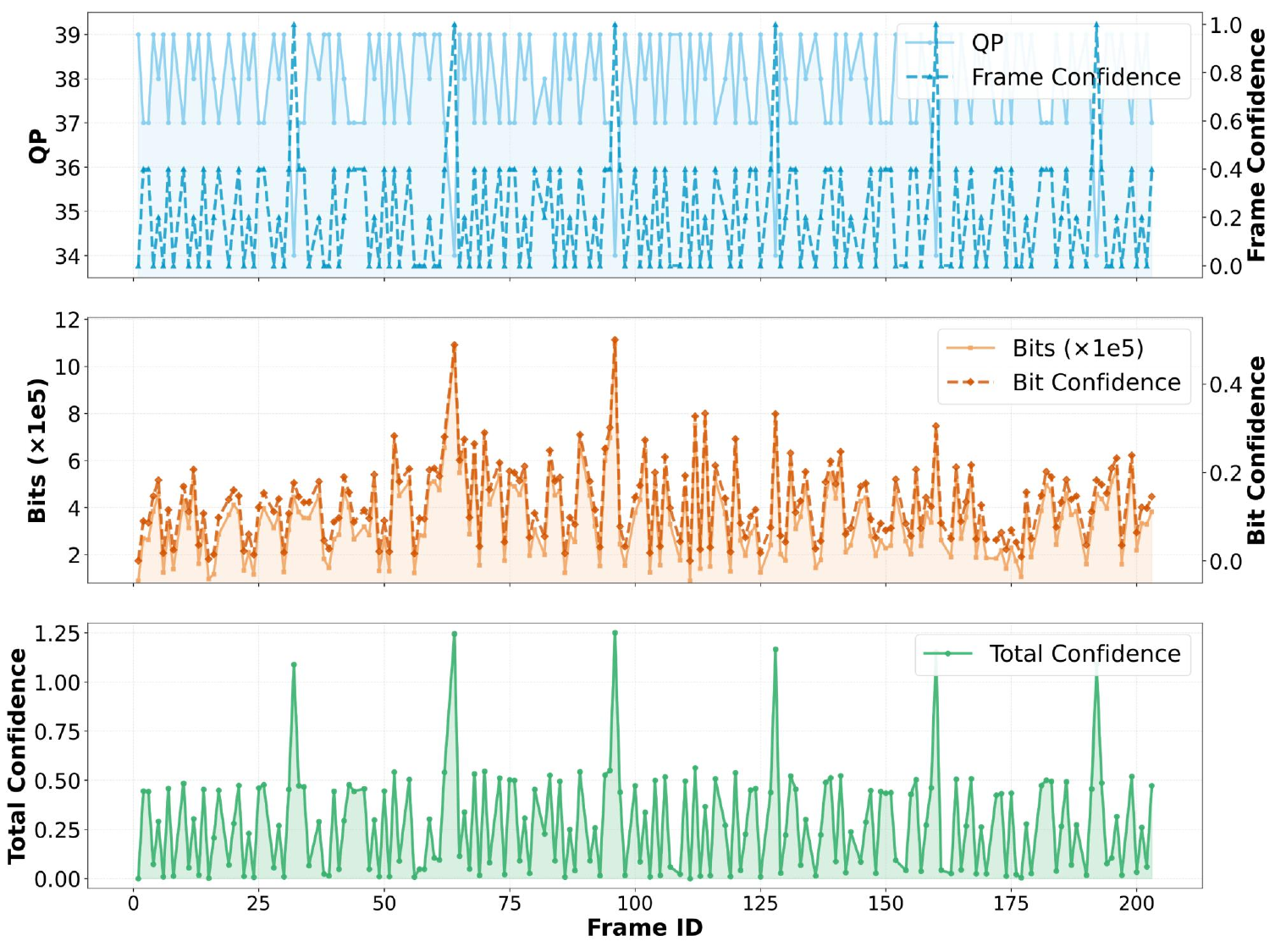}
\caption{\textbf{Frame-wise QP, bitrate, and confidence scores.}
Top: per-frame QP values (blue) and the corresponding QP confidence $q_t^{q}$ (red). 
Middle: per-frame bit allocation (orange) and the derived bitrate confidence $q_t^{b}$ (blue). 
Bottom: combined confidence $q_t$ across the sequence. High-quality frames exhibit higher confidence values, while heavily compressed frames 
show consistently lower confidence.}
\label{fig:supp-fig10}
\end{figure}

\begin{figure}[tbh]
\centering
\includegraphics[width=\columnwidth]{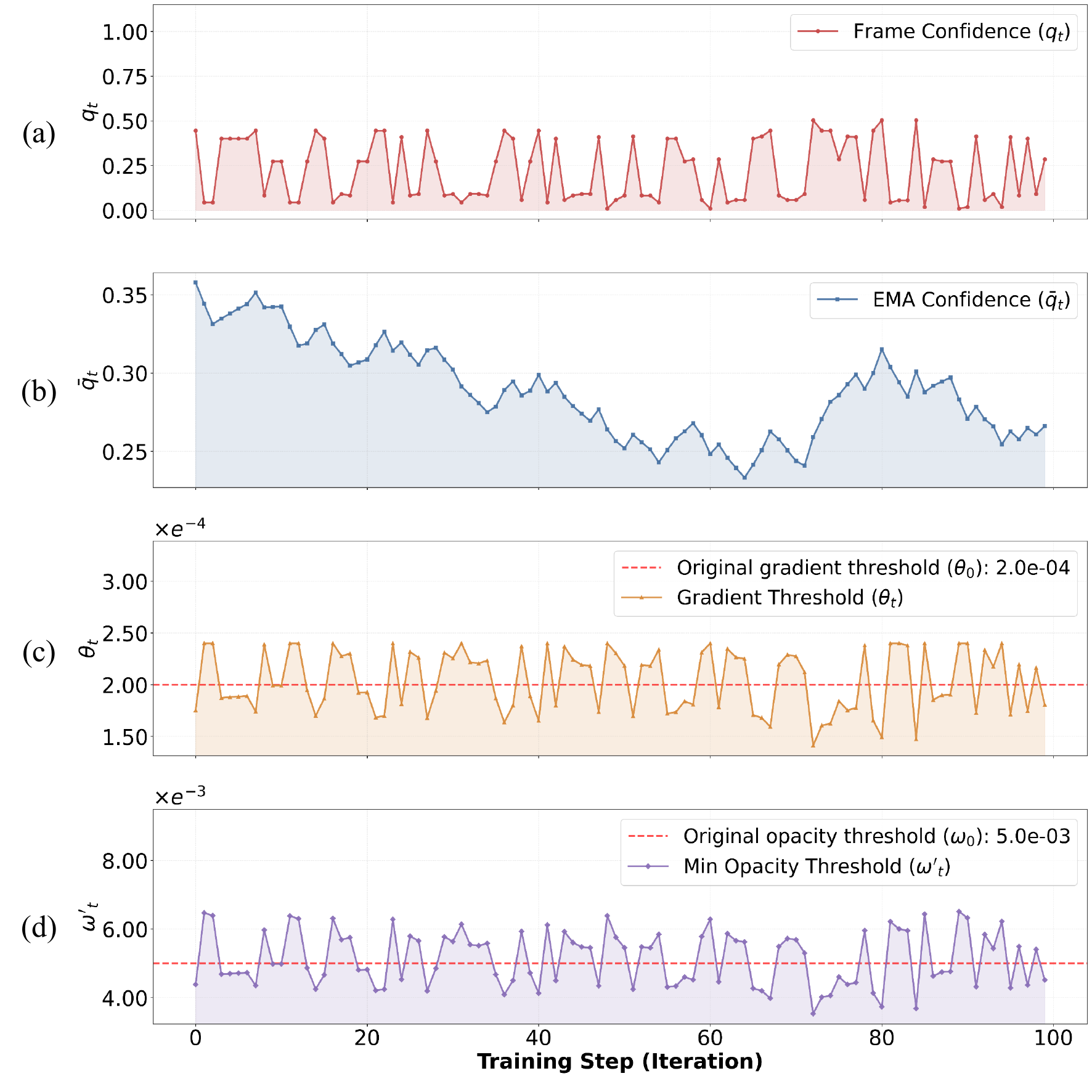}
\caption{\textbf{Analysis of adaptive threshold adjustment over training steps.}
Each plot shows how the corresponding quantity evolves across the first 100 
training iterations: 
(a) frame confidence $q_t$, 
(b) EMA-smoothed confidence $\bar{q}_t$, 
(c) dynamic gradient threshold $\theta_t$, and 
(d) dynamic opacity threshold $\omega'_t$. }
\label{fig:supp-fig11}
\end{figure}

\paragraph{Analysis of QP and Bit Confidence Computation.}
Fig.~\ref{fig:supp-fig10} visualizes how QP and bit allocation fluctuate across the video sequence. The top and middle plots show that both QP values and bitrate vary substantially frame-by-frame, and their corresponding confidence scores ($q_t^{q}$, $q_t^{b}$) follow these trends. This behavior aligns with typical codec operations, where high-QP or low-bit frames lose more high-frequency information, while low-QP or high-bit frames retain richer visual detail. The bottom plot shows the combined confidence $q_t = q_t^{q} + q_t^{b}$, which captures overall frame quality more reliably than either signal alone. High-quality frames consistently yield higher confidence scores, whereas strongly compressed frames produce lower scores, demonstrating that the unified confidence measure effectively reflects frame-wise quality fluctuations.

\paragraph{Analysis of Adaptive Threshold Adjustment.}
Fig.~\ref{fig:supp-fig11} visualizes how our adaptive density control responds to frame-level quality variations during training. In (a), the frame confidence $q_t$ fluctuates according to the compression quality of each frame, while (b) shows its EMA-smoothed form $\bar{q}_t$, which captures the overall quality trend of the sequence. This smoothed baseline acts as a stable reference for modulating the strength of threshold adjustments. Plots (c) and (d) show the dynamic gradient threshold $\theta_t$ and dynamic opacity threshold $\omega'_t$ relative to their fixed originals (red lines). Both thresholds shift in accordance with the confidence signals, lowering or raising their values based on frame reliability.

\begin{figure}[tbh]
\centering
\includegraphics[width=\columnwidth]{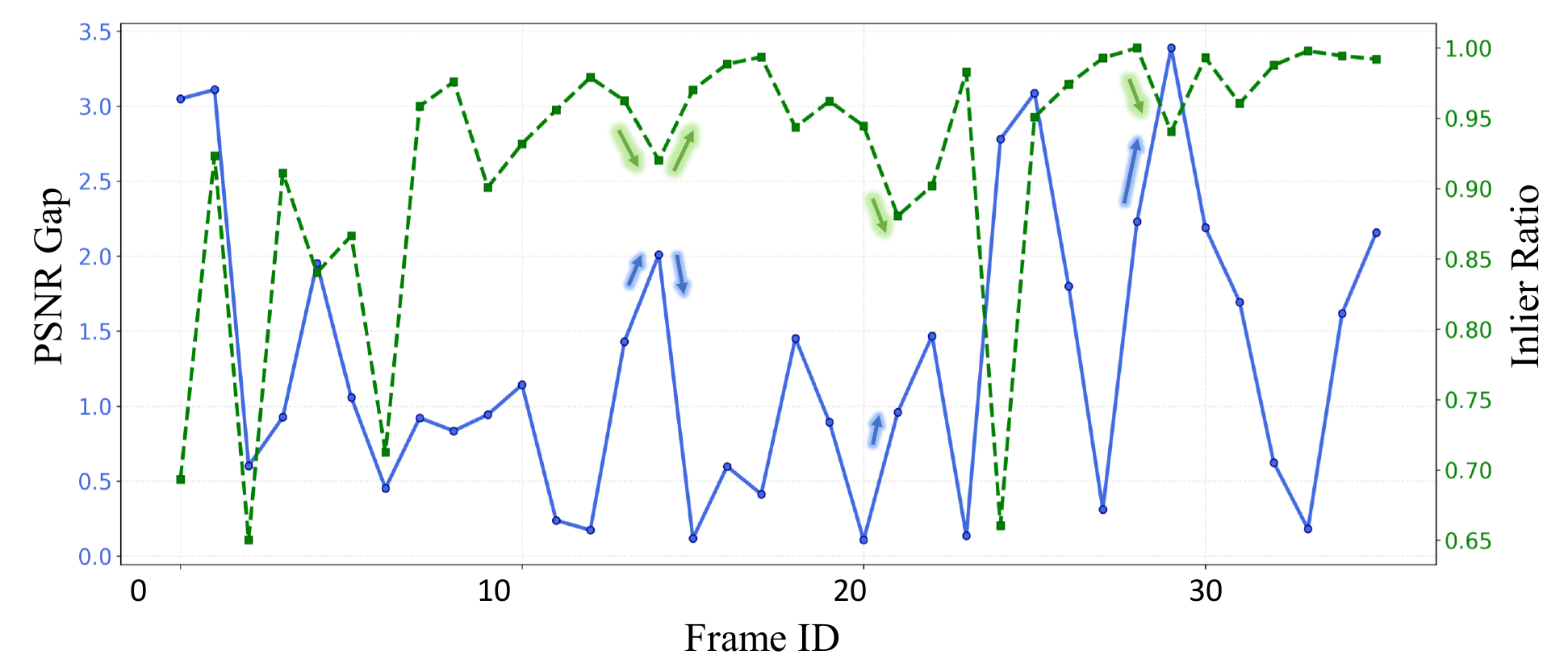}
\caption{\textbf{Correlation between PSNR gap and inlier ratio.}
Larger frame-to-frame PSNR gaps (blue) coincide with drops in the MASt3R \cite{mast3r} inlier 
ratio (green), indicating that quality gaps directly deteriorate feature matching 
reliability.}
\label{fig:supp-fig12}
\end{figure}

\begin{figure}[tbh]
\centering
\includegraphics[width=\columnwidth]{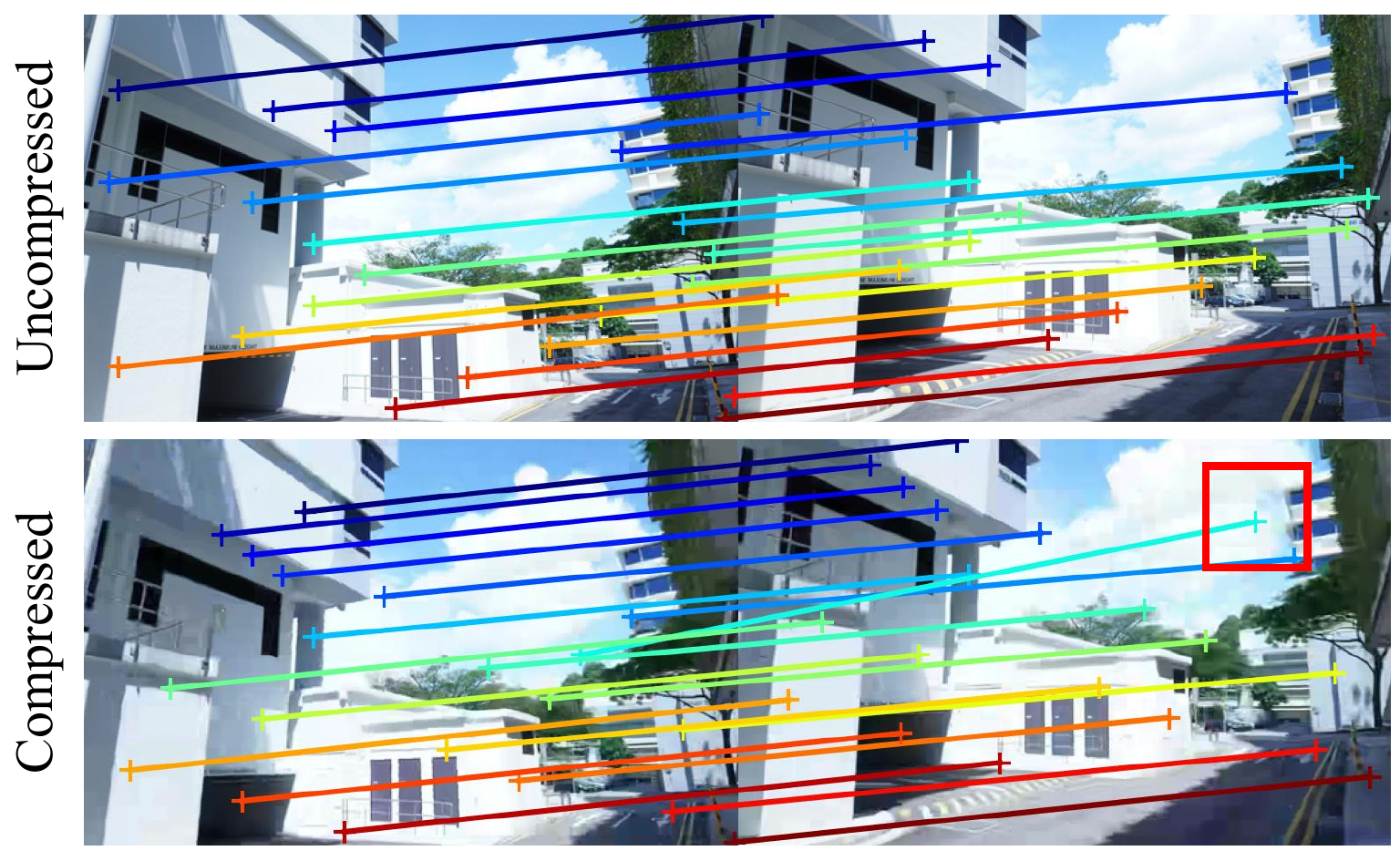}
\caption{\textbf{MASt3R~\cite{mast3r} matching under compression.}
Feature correspondences become noticeably unstable when the input is compressed.
The effect is most apparent in flat, low-texture regions (e.g., sky areas), where compression artifacts disrupt reliable matching and lead to incorrect alignments.
}
\label{fig:supp-fig13}
\end{figure}

\paragraph{Analysis of PSNR Gap Impact on Feature Matching.}
Fig.~\ref{fig:supp-fig12} illustrates the relationship between frame-wise PSNR gaps
and the inlier ratio from MASt3R~\cite{mast3r}. 
When large PSNR gaps occur between adjacent frames, the inlier ratio consistently drops,
indicating degraded feature matching. 
This observation directly supports the second problem discussed in Sec.~3 of the main paper,
where abrupt quality differences undermine pose initialization and destabilize
view supervision. 
We also observe that such quality gaps negatively affect MASt3R correspondences themselves, as shown in Fig.~\ref{fig:supp-fig13}, where compression reduces matching stability.

\section{Additional Ablations}
All additional ablation experiments in this section are evaluated on the Free dataset~\cite{free}, with results averaged over seven scenes.


\begin{table}[t]
\centering
\footnotesize
\setlength{\tabcolsep}{2pt}
\renewcommand{\arraystretch}{0.95}
\caption{\textbf{Comparison of training efficiency on the Free dataset.} All experiments were conducted on a single NVIDIA RTX 3080 GPU.}
\label{tab:supp-tab9}

\begin{tabular}{l|c c c}
\toprule
Method & FPS$\uparrow$ & Time (min)$\downarrow$ & Size (MB)$\downarrow$ \\
\midrule
LongSplat~\cite{longsplat} & \textbf{40.89} & 117.07 & 82.46 \\
Ours                      & 40.86          & \textbf{116.62} & \textbf{82.44} \\
\bottomrule
\end{tabular}
\end{table}

\begin{table}[t]
\centering
\small
\setlength{\tabcolsep}{4pt}
\renewcommand{\arraystretch}{1.12}
\caption{\textbf{Comparison of 2D refinement pipelines.}
We compare representative 2D and 3D refinement approaches in terms of rendering quality and runtime efficiency.}
\label{tab:supp-tab10}

\resizebox{\columnwidth}{!}{%
\begin{tabular}{l|l|cccc}
\toprule
\textbf{Pipeline} & \textbf{Method} 
& PSNR$\uparrow$ & SSIM$\uparrow$ & LPIPS$\downarrow$ & Runtime$\downarrow$ \\
\midrule
Deblurring
& HQGS ~\cite{linhqgs}
& 16.62 & 0.51 & 0.71 & \textbf{17m} \\

2D refinement
& PiSA-SR ~\cite{sun2025pixel} + LongSplat
& 24.53 & 0.71 & 0.33 & 127m \\

2D refinement
& CODiff ~\cite{guo2025compression} + LongSplat
& 23.52 & 0.67 & 0.37 & 149m \\

\midrule
3D
& \textbf{Ours}
& \textbf{25.31} & \textbf{0.74} & \textbf{0.29} & 117m \\
\bottomrule
\end{tabular}%
}
\end{table}

\begin{table}[t]
\centering
\small
\setlength{\tabcolsep}{4pt}
\renewcommand{\arraystretch}{1.1}
\caption{\textbf{Ablation study on confidence score formulations.}
We compare alternative scoring formulations and evaluate their impact on both photometric quality and pose accuracy.}
\label{tab:supp-tab11}
\resizebox{\columnwidth}{!}{%
\begin{tabular}{l|c c c c c c}
\toprule
Method
& PSNR$\uparrow$ & SSIM$\uparrow$ & LPIPS$\downarrow$
& RPE$_t\downarrow$ & RPE$_r\downarrow$ & ATE$\downarrow$ \\
\midrule

Non-linear (\cref{eq:4})
& 24.89 & 0.73 & 0.30
& 0.667 & 1.616 & 0.011 \\

Non-exponential (\cref{eq:9})
& 24.94 & \textbf{0.74} & \textbf{0.29}
& 0.672 & 1.483 & 0.013 \\
\textbf{Ours}
& \textbf{25.31} & \textbf{0.74} & \textbf{0.29}
& \textbf{0.539} & \textbf{1.047} & \textbf{0.008} \\
\bottomrule
\end{tabular}%
}
\end{table}

\paragraph{Training Efficiency.}
\cref{tab:supp-tab9} shows that our method incurs a slight reduction in FPS, while maintaining comparable training time and model size.


\paragraph{Comparison with 2D Refinement-Based Pipelines.} \cref{tab:supp-tab10} compares 2D and 3D refinement approaches on compressed video inputs. Existing 2D pipelines improve compressed frames using diffusion-based models such as PiSA-SR~\cite{sun2025pixel} and CODiff~\cite{guo2025compression}, and then apply LongSplat for reconstruction, resulting in a preprocessing strategy that is decoupled from 3D optimization. In contrast, our method directly performs compression-aware optimization in the 3D Gaussian domain and is specifically tailored for compressed videos. HQGS achieves faster runtime but assumes accurate camera poses and targets image-level degradations such as static JPEG compression, making it unsuitable for long unposed video sequences under real-world video compression, while diffusion-based 2D refinement methods incur high computational cost. These results demonstrate the benefit of direct 3D optimization over 2D refinement-based pipelines.

\paragraph{Ablation on Alternative Scoring Formulations.} We investigate alternative scoring formulations to assess the impact of the functional design of frame confidence. Specifically, we evaluate a non-linear variant that replaces the linear mapping in ~\cref{eq:4} with a sigmoid-based formulation and a non-exponential variant that removes the exponential mapping in ~\cref{eq:9}. For the non-linear variant, we apply a sigmoid-based scoring function to both the QP- and bit-based confidence terms. A temperature parameter $\tau$ controls the smoothness of the score transition, while a scaling factor $\rho$ restricts the contribution of the bit-based term; $\sigma$ denotes the sigmoid function. The resulting frame confidence is defined as
\begin{equation}
q_t=\lambda^q \sigma\!\left((\tilde q_t-0.5)/\tau\right)
+\lambda^b \rho \, \sigma\!\left((\tilde b_t-0.5)/\tau\right).
\end{equation}
For the non-exponential variant, we change the exponential confidence modulation in \cref{eq:9} to a linear adjustment based on the confidence difference between the EMA baseline and the current frame. Specifically, the densification and pruning thresholds are modulated proportionally to the confidence gap $(\bar{q}_t - q_t)$, rather than being scaled exponentially. This modification is formulated as:

\begin{equation}
\theta_t = \theta_0 \left(1 + \alpha (\bar{q}_t - q_t)\right), \qquad
\omega'_t = \omega_0 \left(1 + \alpha (\bar{q}_t - q_t)\right),
\end{equation}
where $\alpha$ is a scaling factor that controls the sensitivity of the threshold modulation to the confidence gap $(\bar{q}_t - q_t)$.

As shown in \cref{tab:supp-tab11}, our default formulation with linear QP–bitrate aggregation and exponential mapping consistently outperforms alternative designs. Directly reflecting QP and bitrate in a linear manner provides a more faithful measure of frame reliability. In addition, the exponential mapping in our formulation smoothly amplifies confidence differences, whereas removing or altering this property can introduce abrupt or unstable confidence changes, which are detrimental to pose optimization.


\begin{table}[t]
\centering
\small
\setlength{\tabcolsep}{3pt}
\renewcommand{\arraystretch}{1.12}
\caption{\textbf{Ablation across different compression levels.} 
We compare LongSplat and our method under QP settings of 27, 37, and 47 on the 
Free dataset, enabling evaluation of performance changes across varying degrees 
of compression.}
\label{tab:supp-tab12}
\resizebox{\columnwidth}{!}{%
\begin{tabular}{l|c|cccccc}
\toprule
\textbf{Method} & \textbf{QP} &
PSNR$\uparrow$ & SSIM$\uparrow$ & LPIPS$\downarrow$ &
RPE$_t\downarrow$ & RPE$_r\downarrow$ & ATE$\downarrow$ \\
\midrule
LongSplat~\cite{longsplat} & \multirow{2}{*}{27}
& 25.54 & \textbf{0.75} & \textbf{0.30} & \textbf{0.631} & 1.426 & \textbf{0.011} \\
\textbf{Ours}              &
& \textbf{25.56} & \textbf{0.75} & \textbf{0.30} & 0.742 & \textbf{1.405} & 0.012 \\
\midrule
LongSplat~\cite{longsplat} & \multirow{2}{*}{37}
& 24.69 & 0.73 & 0.30 & 0.872 & 1.721 & 0.016 \\
\textbf{Ours}              &
& \textbf{25.31} & \textbf{0.74} & \textbf{0.29} & \textbf{0.539} & \textbf{1.047} & \textbf{0.008} \\
\midrule
LongSplat~\cite{longsplat} & \multirow{2}{*}{47}
& 23.72 & \textbf{0.64} & \textbf{0.44} & \textbf{0.842} & \textbf{1.829} & \textbf{0.013} \\
\textbf{Ours}              &
& \textbf{23.80} & \textbf{0.64} & \textbf{0.44} & 0.963 & 1.958 & 0.017 \\
\bottomrule
\end{tabular}%
}
\end{table}

\begin{figure}
\centering
\includegraphics[width=\columnwidth]{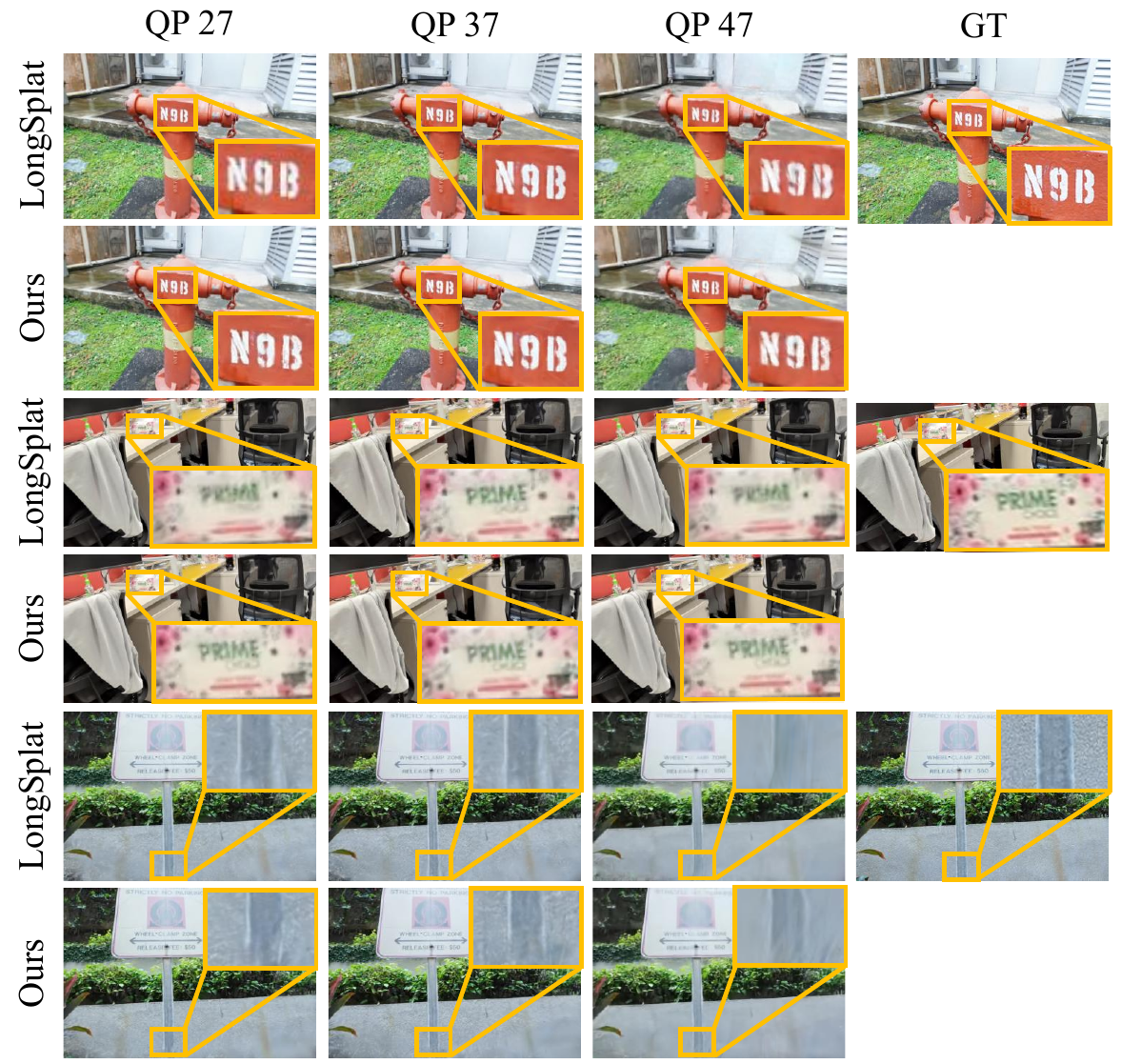}
\caption{\textbf{Rendering comparison between LongSplat and our method across different input video compression levels (QP 27, 37, 47)}. As QP increases, the input frames become more heavily compressed, leading to a noticeable decline in rendering quality for both methods. However, LongSplat exhibits stronger degradation, especially in text regions and fine details, which often appear blurred or broken. In contrast, our method adaptively grows or prunes Gaussians based on compression cues, enabling more faithful preservation of fine structures and clearer boundary reconstruction.}
\label{fig:supp-fig14}
\end{figure}


\paragraph{Effect of Compression Levels.}
\cref{tab:supp-tab12} presents the performance of LongSplat~\cite{longsplat} and our method under different QP 
settings. Across all compression levels (QP 27, 37, and 47), our approach maintains stable reconstruction quality and pose accuracy, demonstrating robustness to varying degrees of compression applied to the Free dataset~\cite{free}. Fig.~\ref{fig:supp-fig14} further illustrates this trend, showing that our method preserves finer textures and clearer structures than LongSplat, especially as compression becomes stronger.


\paragraph{Evaluation with Compressed Video.}
Since uncompressed videos are unavailable in real-world scenarios, \cref{tab:supp-tab13} reports results where all methods are evaluated on the encoded input sequences. Our method shows consistently higher performance across scenes compared to LongSplat.

\begin{table}[t]
\centering
\small
\setlength{\tabcolsep}{3pt}
\renewcommand{\arraystretch}{1.12}
\caption{\textbf{Quantitative comparison on the Free dataset using compressed input frames.}
All methods are evaluated directly on codec-encoded sequences to reflect real-world conditions.}
\label{tab:supp-tab13}
\resizebox{\columnwidth}{!}{%
\begin{tabular}{l!{\vrule width 0.6pt}cccccc|cccccc}
\toprule
\multirow{2}{*}{Scenes} &
\multicolumn{6}{c|}{LongSplat~\cite{longsplat}} &
\multicolumn{6}{c}{\textbf{Ours}} \\
\cmidrule(lr){2-7} \cmidrule(lr){8-13}
& PSNR$\uparrow$ & SSIM$\uparrow$ & LPIPS$\downarrow$ & RPE$_{t}\downarrow$ & RPE$_{r}\downarrow$ & ATE$\downarrow$
& PSNR$\uparrow$ & SSIM$\uparrow$ & LPIPS$\downarrow$ & RPE$_{t}\downarrow$ & RPE$_{r}\downarrow$ & ATE$\downarrow$ \\
\midrule
Grass   & 24.51 & 0.67 & 0.34 & \textbf{0.092} & \textbf{1.270} & \textbf{0.006} & \textbf{24.69} & \textbf{0.68} & \textbf{0.33} & 0.322 & 2.115 & 0.007 \\
Hydrant & \textbf{24.31} & \textbf{0.77} & \textbf{0.25} & \textbf{0.815} & \textbf{0.838} & \textbf{0.022} & 23.96 & 0.75 & 0.26 & 0.963 & 1.306 & 0.031 \\
Lab     & \textbf{26.09} & \textbf{0.85} & \textbf{0.18} & \textbf{1.230} & 1.499 & \textbf{0.016} & 26.06 & \textbf{0.85} & \textbf{0.18} & 1.249 & \textbf{1.497} & \textbf{0.016} \\
Pillar  & 27.37 & 0.79 & 0.26 & 0.159 & 0.272 & 0.003 & \textbf{28.43} & \textbf{0.81} & \textbf{0.25} & \textbf{0.043} & \textbf{0.142} & \textbf{0.002} \\
Road    & 21.59 & \textbf{0.71} & \textbf{0.35} & 2.660 & 5.882 & 0.035 & \textbf{21.67} & \textbf{0.71} & \textbf{0.35} & \textbf{2.122} & \textbf{3.524} & \textbf{0.027} \\
Sky     & 25.84 & 0.85 & \textbf{0.20} & 0.467 & \textbf{1.521} & 0.011 & \textbf{26.30} & \textbf{0.86} & \textbf{0.20} & \textbf{0.364} & \textbf{1.521} & \textbf{0.007} \\
Stair   & 27.46 & \textbf{0.83} & 0.24 & 0.135 & 0.205 & 0.003 & \textbf{27.77} & \textbf{0.83} & \textbf{0.23} & \textbf{0.134} & \textbf{0.165} & \textbf{0.002} \\
\midrule
Avg     & 25.31 & \textbf{0.78} & \textbf{0.26} & 0.794 & 1.641 & 0.014 & \textbf{25.55} & \textbf{0.78} & \textbf{0.26} & \textbf{0.742} & \textbf{1.467} & \textbf{0.013} \\
\bottomrule
\end{tabular}%
}
\end{table}

\paragraph{Ablation on Parameter $\lambda^q$ and $\lambda^b$.}
\cref{tab:supp-tab14} analyzes the effect of the weighting coefficients $\lambda^{q}$ and $\lambda^{b}$. Assigning a larger weight to the QP-based term consistently yields better photometric and pose accuracy, as QP directly reflects compression quality and frame reliability, whereas bitrate can vary due to scene complexity. Using a large bitrate weight degrades pose accuracy, and balanced weights lead to suboptimal performance. Accordingly, we adopt $\lambda^{q}=1.0$ and $\lambda^{b}=0.5$, which achieves the best overall performance.

\paragraph{Ablation on Momentum Parameter $\beta$.}
\cref{tab:supp-tab15} reports the effect of the momentum parameter $\beta$. When $\beta$ is set to 0.9, the updates become relatively unstable, leading to lower reconstruction quality across PSNR, SSIM, and LPIPS. In contrast, assigning a very high value such as 0.99 results in excessive momentum accumulation, making the model less responsive to changes and degrading both reconstruction metrics and pose-related errors (RPE and ATE). When the momentum parameter is set to 0.95, the model achieves the best overall performance, providing a stable balance between responsiveness and smoothness. These findings suggest that choosing a well-balanced momentum value is essential for achieving stable updates and reliable performance throughout the pipeline.

\paragraph{Ablation on Quality-Gap Parameter $\eta$.}
\cref{tab:supp-tab16} evaluates the effect of the pixel-drop scaling
parameter $\eta$, which determines how strongly the inlier-ratio gap influences
the masking applied to photometric supervision. Increasing $\eta$ suppresses a
larger portion of unreliable pixels, while smaller values reduce this effect.
Among the tested settings, $\eta=0.5$ provides the best overall performance.

\begin{table}[t]
\centering
\small
\setlength{\tabcolsep}{4pt}
\renewcommand{\arraystretch}{1.1}
\caption{\textbf{Ablation on weighting coefficients $\lambda^{q}$ and $\lambda^{b}$ for frame-quality estimation.}
We study how the QP-based score weight ($\lambda^{q}$) and the bitrate-confidence score weight ($\lambda^{b}$) affect both photometric quality  and pose accuracy.}
\label{tab:supp-tab14}

\resizebox{\columnwidth}{!}{%
\begin{tabular}{c c | c c c c c c}
\toprule
$\lambda^{q}$ & $\lambda^{b}$
& PSNR$\uparrow$ & SSIM$\uparrow$ & LPIPS$\downarrow$
& RPE$_t\downarrow$ & RPE$_r\downarrow$ & ATE$\downarrow$ \\
\midrule
0.5 & 0.5
& 24.96 & 0.74 & 0.30
& 0.875 & 1.353 & 0.015 \\

0.5 & 1.0
& 24.81 & 0.73 & 0.30
& 0.743 & 1.589 & 0.013 \\

\textbf{1.0} & \textbf{0.5}
& \textbf{25.31} & \textbf{0.74} & \textbf{0.29}
& \textbf{0.539} & \textbf{1.047} & \textbf{0.008} \\

1.0 & 1.0
& 24.94 & 0.74 & 0.29
& 0.822 & 1.705 & 0.013 \\
\bottomrule
\end{tabular}%
}
\end{table}

\begin{table}[t]
\centering
\small
\setlength{\tabcolsep}{3pt}
\renewcommand{\arraystretch}{1.12}
\caption{\textbf{Ablation on momentum parameter $\beta$ evaluated on the Free dataset.}
Results are shown for different EMA momentum values, with the quality-gap parameter
$\eta$ fixed at 0.5.}
\label{tab:supp-tab15}

\resizebox{\columnwidth}{!}{%
\begin{tabular}{l|cccccc}
\toprule
\textbf{$\beta$} &
PSNR$\uparrow$ & SSIM$\uparrow$ & LPIPS$\downarrow$ &
RPE$_t\downarrow$ & RPE$_r\downarrow$ & ATE$\downarrow$ \\
\midrule
0.9
& 24.80 & 0.73 & 0.30 & 0.762 & 2.133 & 0.013 \\

0.95 (\textbf{Ours})
& \textbf{25.31} & \textbf{0.74} & \textbf{0.29}
& \textbf{0.539} & \textbf{1.047} & \textbf{0.008} \\

0.99
& 24.95 & 0.73 & 0.30 & 0.667 & 1.401 & 0.012 \\
\bottomrule
\end{tabular}%
}
\end{table}

\begin{table}[t]
\centering
\small
\setlength{\tabcolsep}{3pt}
\renewcommand{\arraystretch}{1.12}
\caption{\textbf{Ablation on the quality-gap parameter $\eta$.}
Performance comparison on the Free dataset for different values of $\eta$, which controls the strength of quality-gap masking.
All experiments fix the momentum parameter to $\beta=0.95$.}
\label{tab:supp-tab16}

\resizebox{\columnwidth}{!}{%
\begin{tabular}{l|cccccc}
\toprule
\textbf{$\eta$} &
PSNR$\uparrow$ & SSIM$\uparrow$ & LPIPS$\downarrow$ &
RPE$_t\downarrow$ & RPE$_r\downarrow$ & ATE$\downarrow$ \\
\midrule
0.3
& 23.23 & 0.68 & 0.33 & 1.013 & 3.097 & 0.017 \\

0.5 (\textbf{Ours})
& \textbf{25.31} & \textbf{0.74} & \textbf{0.29}
& \textbf{0.539} & \textbf{1.047} & \textbf{0.008} \\

0.7
& 24.81 & 0.73 & 0.30 & 0.817 & 1.755 & 0.014 \\
\bottomrule
\end{tabular}%
}
\end{table}





\section{Dataset Construction Details: Video Compression Settings}

\begin{table*}[t]
\small
\caption{\textbf{Encoding configuration.} FFmpeg command used for x265 encoding.}
\label{tab:supp-tab17}
\centering
\renewcommand{\arraystretch}{1.05}

\begin{tabular}{p{0.97\textwidth}}
\hline
\texttt{ffmpeg -hide\_banner -y -f concat -safe 0 -r 60 -i input\_list.txt -vsync 0} \\
\texttt{-vf "scale=trunc(iw/2)*2:trunc(ih/2)*2, format=yuv420p, setsar=1"} \\
\texttt{-pix\_fmt yuv420p -c:v libx265 -preset medium -tune psnr -profile:v main -g 32} \\
\texttt{-x265-params "qp=27:keyint=32:min-keyint=1:scenecut=40:open-gop=0:} \\
\texttt{csv=logs/:sequence\_x265\_qp27.csv:csv-log-level=1:psnr=1:ssim=1:aq-mode=0"} \\
\texttt{outputs/sequence\_x265\_qp27.mp4 > logs/sequence\_x265\_qp27\_encode.log 2>\&1} \\
\hline
\end{tabular}

\end{table*}




We provide more detailed explanations of the video compression settings used in our experiments.

\paragraph{Encoder Configuration.} For all image sequences used in the main paper, including the Tanks and Temples 
and the Free datasets, we encoded the input frames using \texttt{ffmpeg} and the \texttt{libx265} at 60~fps. We employ the HEVC encoder \texttt{libx265} (\texttt{-c:v libx265}) with the \texttt{medium} preset, \texttt{-tune psnr}, and the HEVC Main profile (\texttt{-profile:v main}).
The maximum GOP length is set to 32 frames via \texttt{-g 32}.
Additional encoder parameters are passed through \texttt{-x265-params}: 
\texttt{qp} enables constant-QP mode (we use several QP values, e.g., 27/37/47), 
\texttt{keyint=32} and \texttt{min-keyint=1} control the maximum and minimum I-frame interval, 
\texttt{scenecut=40} activates moderate scene-cut detection, 
and \texttt{open-gop=0} disables open GOPs to obtain a simpler temporal structure.

\subsubsection*{Encoding Configuration}

Various quality factors can be used by setting the QP value.
For clarity and reproducibility, we provide the exact FFmpeg command used to produce the QP\,=\,27 versions of the dataset. This command specifies consistent codec parameters such as GOP structure, chroma format, and logging full configuration is summarized in \cref{tab:supp-tab17}.

\paragraph{Logging and Quality Metrics.}
For each encoding, x265 writes per-frame statistics to a CSV file using 
\texttt{csv=...} and \texttt{csv-log-level=1}; these logs include the frame type, 
frame-level QP, number of bits, POC, encode order, and frame-level quality metrics 
(PSNR and SSIM for Y, U, V, and YUV channels). Adaptive quantization is disabled 
(\texttt{aq-mode=0}) so that the effective quantization closely follows the nominal QP.
In addition to frame-level quantities, the CSV logs produced by x265 
(\texttt{csv-log-level=1}) include several block-level coding statistics aggregated 
over all Coding Tree Units (CTUs) in each frame. These include the percentage of 
skip blocks and merge-mode blocks at multiple block sizes (64$\times$64, 32$\times$32, 
16$\times$16, 8$\times$8), as well as the GOP position and temporal-layer index. 
These statistics provide a compact summary of the encoder’s intra/inter mode 
decisions and temporal prediction structure. Finally, all console output from 
ffmpeg (stdout and stderr) is redirected into a single log file for reproducibility.

\section{Complete Quantitative Evaluation}

\begin{table*}[tbh]
\centering
\setlength{\tabcolsep}{0.8pt}
\caption{\textbf{Quantitative evaluation of camera pose estimation on the Free dataset~\cite{free}.}
We apply our method to both CF-3DGS~\cite{cf3dgs} and LongSplat~\cite{longsplat}. CF-3DGS fails to estimate a valid camera trajectory due to OOM when processing long sequences. Our method significantly reduces pose estimation errors for LongSplat across all scenes.}
\label{tab:supp-tab18}

\resizebox{\textwidth}{!}{%
\begin{tabular}{l!{\vrule width 0.6pt}ccc|ccc|ccc|ccc|ccc|ccc}
\toprule
\multirow{2}{*}{\textbf{Scene}} &
\multicolumn{3}{c|}{\textcolor{gray}{\textbf{LongSplat (Uncomp.)}}} &
\multicolumn{3}{c|}{\textbf{NoPe-NeRF~\cite{nopenerf}}} &
\multicolumn{3}{c|}{\textbf{CF-3DGS~\cite{cf3dgs}}} &
\multicolumn{3}{c|}{\textbf{CF-3DGS + Ours}} &
\multicolumn{3}{c|}{\textbf{LongSplat~\cite{longsplat}}} &
\multicolumn{3}{c}{\textbf{LongSplat + Ours}} \\
\cmidrule(lr){2-4} \cmidrule(lr){5-7} \cmidrule(lr){8-10}
\cmidrule(lr){11-13} \cmidrule(lr){14-16} \cmidrule(lr){17-19}
& RPE$_t\downarrow$ & RPE$_r\downarrow$ & ATE$\downarrow$
& RPE$_t\downarrow$ & RPE$_r\downarrow$ & ATE$\downarrow$
& RPE$_t\downarrow$ & RPE$_r\downarrow$ & ATE$\downarrow$
& RPE$_t\downarrow$ & RPE$_r\downarrow$ & ATE$\downarrow$
& RPE$_t\downarrow$ & RPE$_r\downarrow$ & ATE$\downarrow$
& RPE$_t\downarrow$ & RPE$_r\downarrow$ & ATE$\downarrow$ \\
\midrule

Grass
& \textcolor{gray}{0.037} & \textcolor{gray}{0.255} & \textcolor{gray}{0.001}
& 3.631 & 10.256 & 0.353
& OOM & OOM & OOM
& OOM & OOM & OOM
& 0.092 & 1.270 & 0.006
& 0.061 & 0.550 & 0.002 \\

Hydrant
& \textcolor{gray}{1.161} & \textcolor{gray}{1.419} & \textcolor{gray}{0.029}
& 5.198 & 4.990 & 0.523
& 1.163 & 7.550 & 0.042
& 1.161 & 7.908 & 0.043
& 1.101 & 1.155 & 0.034
& 0.044 & 0.367 & 0.002 \\

Lab
& \textcolor{gray}{1.161} & \textcolor{gray}{2.163} & \textcolor{gray}{0.013}
& 6.186 & 2.597 & 0.489
& OOM & OOM & OOM
& OOM & OOM & OOM
& 1.491 & 1.744 & 0.023
& 1.249 & 1.497 & 0.016 \\

Pillar
& \textcolor{gray}{0.057} & \textcolor{gray}{0.394} & \textcolor{gray}{0.002}
& 4.355 & 4.943 & 0.518
& 0.656 & 5.971 & 0.033
& 0.704 & 6.164 & 0.026
& 0.159 & 0.272 & 0.003
& 0.043 & 0.142 & 0.002 \\

Road
& \textcolor{gray}{2.018} & \textcolor{gray}{5.738} & \textcolor{gray}{0.026}
& 6.511 & 3.655 & 0.433
& OOM & OOM & OOM
& OOM & OOM & OOM
& 2.660 & 5.882 & 0.035
& 1.885 & 3.138 & 0.028 \\

Sky
& \textcolor{gray}{0.361} & \textcolor{gray}{1.425} & \textcolor{gray}{0.008}
& 8.169 & 6.419 & 0.848
& OOM & OOM & OOM
& OOM & OOM & OOM
& 0.467 & 1.521 & 0.011
& 0.364 & 1.521 & 0.007 \\

Stair
& \textcolor{gray}{0.137} & \textcolor{gray}{0.187} & \textcolor{gray}{0.002}
& 6.791 & 0.156 & 0.656
& 0.602 & 5.203 & 0.016
& 0.607 & 6.057 & 0.019
& 0.135 & 0.205 & 0.003
& 0.125 & 0.115 & 0.002 \\

\midrule
Avg.
& \textcolor{gray}{0.705} & \textcolor{gray}{1.654} & \textcolor{gray}{0.012}
& 5.834 & 4.717 & 0.546
& 0.807 & 6.241 & 0.030
& 0.824 & 6.710 & 0.029
& 0.872 & 1.721 & 0.016
& \textbf{0.539} & \textbf{1.047} & \textbf{0.008} \\
\bottomrule
\end{tabular}%
}
\end{table*}

\begin{table*}[t]
\centering
\setlength{\tabcolsep}{0.8pt} 
\caption{\textbf{Quantitative evaluation of novel view synthesis quality on the Tanks and Temples dataset \cite{tankandtemples}.}
We apply our method to both CF-3DGS \cite{cf3dgs} and LongSplat \cite{longsplat}. Our method achieves improved rendering quality across most scenes. ``Uncomp.'' denotes training and evaluation on original uncompressed videos, while other entries are trained on compressed videos and evaluated against uncompressed videos.}
\label{tab:supp-tab19}

\resizebox{\textwidth}{!}{%
\begin{tabular}{l!{\vrule width 0.6pt}ccc|ccc|ccc|ccc|ccc|ccc}
\toprule
\multirow{2}{*}[-3pt]{\textbf{Scene}} &
\multicolumn{3}{c|}{\textcolor{gray}{\textbf{LongSplat (Uncomp.)}}} &
\multicolumn{3}{c|}{\textbf{NoPe-NeRF~\cite{nopenerf}}} &
\multicolumn{3}{c|}{\textbf{CF-3DGS~\cite{cf3dgs}}} &
\multicolumn{3}{c|}{\textbf{CF-3DGS + Ours}} &
\multicolumn{3}{c|}{\textbf{LongSplat~\cite{longsplat}}} &
\multicolumn{3}{c}{\textbf{LongSplat + Ours}} \\
\cmidrule(lr){2-4} \cmidrule(lr){5-7} \cmidrule(lr){8-10}
\cmidrule(lr){11-13} \cmidrule(lr){14-16} \cmidrule(lr){17-19}
& PSNR$\uparrow$ & SSIM$\uparrow$ & LPIPS$\downarrow$
& PSNR$\uparrow$ & SSIM$\uparrow$ & LPIPS$\downarrow$
& PSNR$\uparrow$ & SSIM$\uparrow$ & LPIPS$\downarrow$
& PSNR$\uparrow$ & SSIM$\uparrow$ & LPIPS$\downarrow$
& PSNR$\uparrow$ & SSIM$\uparrow$ & LPIPS$\downarrow$
& PSNR$\uparrow$ & SSIM$\uparrow$ & LPIPS$\downarrow$ \\
\midrule

Church
& \textcolor{gray}{32.31} & \textcolor{gray}{0.94} & \textcolor{gray}{0.10}
& 24.01 & 0.70 & 0.44
& 27.32 & 0.83 & 0.26
& 28.39 & 0.83 & 0.31
& 28.84 & 0.85 & 0.25
& 28.92 & 0.85 & 0.25 \\

Barn
& \textcolor{gray}{27.53} & \textcolor{gray}{0.79} & \textcolor{gray}{0.21}
& 28.48 & 0.82 & 0.33
& 25.53 & 0.72 & 0.33
& 27.60 & 0.85 & 0.21
& 27.51 & 0.78 & 0.29
& 28.12 & 0.79 & 0.29 \\

Museum
& \textcolor{gray}{25.00} & \textcolor{gray}{0.78} & \textcolor{gray}{0.14}
& 28.84 & 0.85 & 0.27
& 27.05 & 0.80 & 0.25
& 26.99 & 0.80 & 0.25
& 29.12 & 0.85 & 0.22
& 29.14 & 0.85 & 0.22 \\

Horse
& \textcolor{gray}{27.96} & \textcolor{gray}{0.89} & \textcolor{gray}{0.12}
& 25.52 & 0.82 & 0.32
& 28.65 & 0.86 & 0.22
& 27.43 & 0.83 & 0.26
& 29.45 & 0.87 & 0.21
& 29.48 & 0.87 & 0.21 \\

Ballroom
& \textcolor{gray}{26.85} & \textcolor{gray}{0.88} & \textcolor{gray}{0.15}
& 26.25 & 0.78 & 0.33
& 27.63 & 0.84 & 0.21
& 29.70 & 0.83 & 0.29
& 25.91 & 0.82 & 0.24
& 26.17 & 0.82 & 0.24 \\

Francis
& \textcolor{gray}{31.66} & \textcolor{gray}{0.90} & \textcolor{gray}{0.17}
& 28.17 & 0.83 & 0.40
& 29.48 & 0.82 & 0.29
& 28.98 & 0.86 & 0.22
& 29.98 & 0.82 & 0.29
& 30.04 & 0.83 & 0.29 \\

Ignatius
& \textcolor{gray}{30.26} & \textcolor{gray}{0.91} & \textcolor{gray}{0.11}
& 26.41 & 0.77 & 0.36
& 25.87 & 0.71 & 0.35
& 25.80 & 0.71 & 0.35
& 26.64 & 0.73 & 0.35
& 26.72 & 0.73 & 0.35 \\

\midrule
Avg.
& \textcolor{gray}{28.80} & \textcolor{gray}{0.87} & \textcolor{gray}{0.14}
& 26.81 & 0.80 & 0.35
& 27.36 & 0.80 & 0.27
& 27.84 & 0.82 & 0.27
& 28.21 & 0.82 & 0.26
& 28.37 & 0.82 & 0.26 \\
\bottomrule
\end{tabular}%
}
\end{table*}

\begin{table*}[thb]
\centering
\setlength{\tabcolsep}{0.8pt} 
\caption{\textbf{Quantitative evaluation of camera pose estimation on the Tanks and Temples dataset \cite{tankandtemples}.}
We apply our method to both CF-3DGS \cite{cf3dgs} and LongSplat \cite{longsplat}. Our method achieves improved pose accuracy across most scenes. ``Uncomp.'' denotes training and evaluation on original uncompressed videos, while other entries are trained on compressed videos and evaluated against uncompressed videos.}
\label{tab:supp-tab20}

\resizebox{\textwidth}{!}{%
\begin{tabular}{l!{\vrule width 0.6pt}ccc|ccc|ccc|ccc|ccc|ccc}
\toprule
\multirow{2}{*}[-3pt]{\textbf{Scene}} &
\multicolumn{3}{c|}{\textcolor{gray}{\textbf{LongSplat (Uncomp.)}}} &
\multicolumn{3}{c|}{\textbf{NoPe-NeRF}} &
\multicolumn{3}{c|}{\textbf{CF-3DGS}} &
\multicolumn{3}{c|}{\textbf{CF-3DGS + Ours}} &
\multicolumn{3}{c|}{\textbf{LongSplat}} &
\multicolumn{3}{c}{\textbf{LongSplat + Ours}} \\
\cmidrule(lr){2-4} \cmidrule(lr){5-7} \cmidrule(lr){8-10}
\cmidrule(lr){11-13} \cmidrule(lr){14-16} \cmidrule(lr){17-19}
& RPE$_t\downarrow$ & RPE$_r\downarrow$ & ATE$\downarrow$
& RPE$_t\downarrow$ & RPE$_r\downarrow$ & ATE$\downarrow$
& RPE$_t\downarrow$ & RPE$_r\downarrow$ & ATE$\downarrow$
& RPE$_t\downarrow$ & RPE$_r\downarrow$ & ATE$\downarrow$
& RPE$_t\downarrow$ & RPE$_r\downarrow$ & ATE$\downarrow$
& RPE$_t\downarrow$ & RPE$_r\downarrow$ & ATE$\downarrow$ \\
\midrule

Church
& \textcolor{gray}{0.502} & \textcolor{gray}{0.377} & \textcolor{gray}{0.017}
& 0.246 & 0.070 & 0.112
& 0.013 & 0.041 & 0.003
& 0.030 & 0.044 & 0.003
& 0.536 & 0.376 & 0.018
& 0.506 & 0.379 & 0.016 \\

Barn
& \textcolor{gray}{0.654} & \textcolor{gray}{1.302} & \textcolor{gray}{0.009}
& 0.027 & 0.017 & 0.003
& 0.048 & 0.142 & 0.007
& 0.029 & 0.022 & 0.003
& 0.847 & 0.341 & 0.013
& 0.594 & 0.280 & 0.009 \\

Museum
& \textcolor{gray}{1.809} & \textcolor{gray}{1.995} & \textcolor{gray}{0.010}
& 0.202 & 0.243 & 0.019
& 0.054 & 0.147 & 0.006
& 0.054 & 0.148 & 0.006
& 0.809 & 0.500 & 0.007
& 0.867 & 0.523 & 0.007 \\

Horse
& \textcolor{gray}{1.141} & \textcolor{gray}{1.891} & \textcolor{gray}{0.012}
& 1.245 & 0.135 & 0.030
& 0.129 & 0.093 & 0.005
& 0.013 & 0.041 & 0.003
& 0.992 & 0.715 & 0.010
& 0.875 & 0.686 & 0.009 \\

Ballroom
& \textcolor{gray}{7.549} & \textcolor{gray}{2.212} & \textcolor{gray}{0.035}
& 0.060 & 0.027 & 0.003
& 0.029 & 0.021 & 0.003
& 0.041 & 0.174 & 0.008
& 7.852 & 2.745 & 0.051
& 7.763 & 2.747 & 0.049 \\

Francis
& \textcolor{gray}{1.314} & \textcolor{gray}{0.979} & \textcolor{gray}{0.032}
& 0.534 & 0.213 & 0.035
& 0.041 & 0.174 & 0.008
& 0.129 & 0.093 & 0.005
& 4.435 & 0.881 & 0.108
& 4.227 & 0.877 & 0.103 \\

Ignatius
& \textcolor{gray}{0.530} & \textcolor{gray}{0.354} & \textcolor{gray}{0.012}
& 0.047 & 0.008 & 0.005
& 0.047 & 0.074 & 0.008
& 0.047 & 0.074 & 0.008
& 0.387 & 0.259 & 0.009
& 0.401 & 0.267 & 0.009 \\

\midrule
Avg.
& \textcolor{gray}{1.928} & \textcolor{gray}{1.301} & \textcolor{gray}{0.018}
& 0.337 & 0.102 & 0.030
& 0.052 & 0.099 & 0.006
& 0.049 & 0.085 & 0.005
& 2.265 & 0.831 & 0.031
& 2.176 & 0.823 & 0.029 \\
\bottomrule
\end{tabular}%
}
\end{table*}

\subsection{Free Dataset}
To complement the averaged pose metrics reported in the main paper, 
we provide scene-wise pose estimation results for the Free dataset~\cite{free} in \cref{tab:supp-tab18}. These results show that our method consistently improves both translational and rotational accuracy across all scenes.

\subsection{Tanks and Temples Dataset}
Along with the averaged NVS reconstruction scores reported in the main paper, we include scene-wise NVS quality results for the Tanks and Temples dataset in \cref{tab:supp-tab19}. In addition, \cref{tab:supp-tab20} presents the scene-wise camera pose estimation results, providing a more detailed analysis beyond the averaged metrics. Our method consistently improves both rendering fidelity and pose accuracy across individual scenes, demonstrating stronger robustness to compression-related degradation.

\subsection{Hike Dataset}
To complement the averaged results reported in the main paper, we provide scene-wise novel view synthesis quality and camera pose estimation results on the Hike dataset in \cref{tab:supp-tab21}. Our method consistently improves both photometric quality and pose accuracy across most scenes compared to LongSplat, demonstrating robust performance under long trajectories, complex camera motion, and large-scale outdoor geometry.

\section{Additional Visual Comparisons}
\subsection{Additional Free Dataset Results}
\begin{figure*}[th]
\centering
\includegraphics[width=\textwidth]{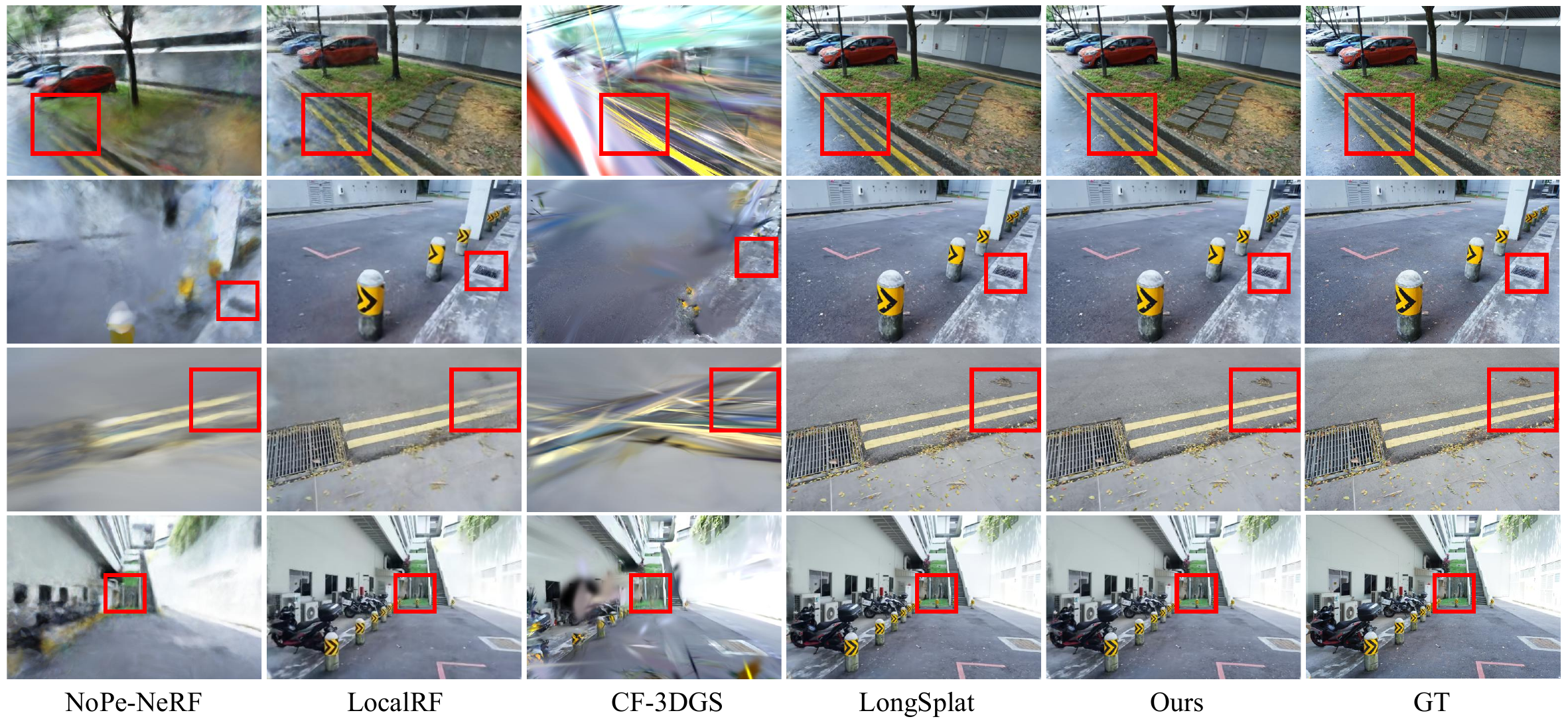}
\caption{\textbf{Further qualitative comparisons on the Free dataset~\cite{free}.} Compared to existing methods such as NoPe-NeRF~\cite{nopenerf}, LocalRF~\cite{localrf}, and CF-3DGS~\cite{cf3dgs}, our approach produces visually clearer and more stable reconstructions under compressed video inputs. Despite the blur and detail loss inherent in compressed frames, our method better preserves structural boundaries, reduces smoothing artifacts, and maintains sharper details across scenes. Furthermore, compared with LongSplat~\cite{longsplat}, our method explicitly considers the quality degradation caused by compression and thus enhances finer details, producing sharper and more expressive structures.}
\label{fig:supp-fig15}
\end{figure*}

Additional qualitative comparisons on the Free dataset~\cite{free} are presented in Fig.~\ref{fig:supp-fig15}. Our method produces clearer object boundaries and alleviates blurring effectively, resulting in sharper and more structurally accurate reconstructions.

\subsection{Additional Tanks and Temples Dataset Results}
\begin{figure*}[th]
\centering
\includegraphics[width=\textwidth]{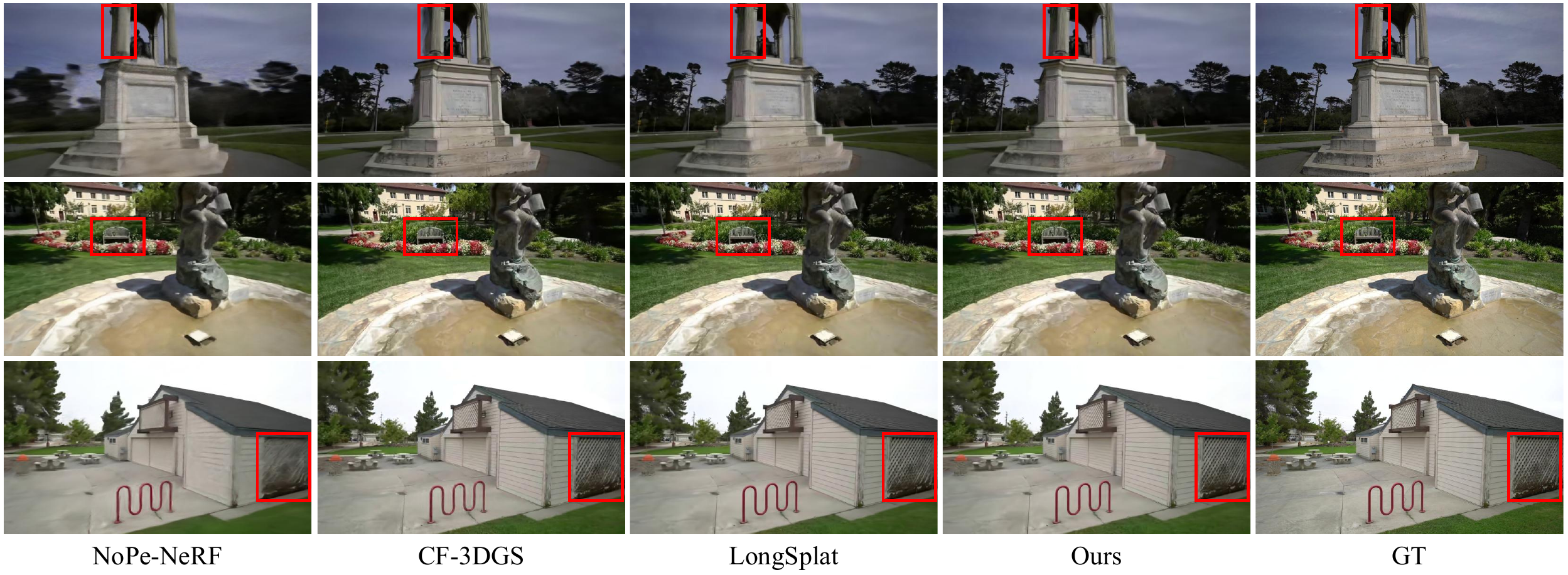}
\caption{\textbf{Further qualitative comparisons on the Tanks and Temples dataset~\cite{tankandtemples}.} Compared with existing methods such as NoPe-NeRF~\cite{nopenerf}, CF-3DGS~\cite{cf3dgs}, and LongSplat~\cite{longsplat}, our method produces sharper and more stable reconstructions under compressed video inputs. Across diverse scenes, it better preserves structural boundaries and fine geometric details, mitigating the artifacts introduced by compression and resulting in more consistent and expressive 3D structures.}
\label{fig:supp-fig16}
\end{figure*}

Additional qualitative comparisons on the Tanks and Temples dataset~\cite{tankandtemples} are shown in Fig.~\ref{fig:supp-fig16}. Our method captures fine-grained details more faithfully and tends to mitigate blurring, resulting in more stable and detailed geometry.

\subsection{Additional Hike Dataset Results}
\begin{figure*}[thb]
\centering
\includegraphics[width=\textwidth]{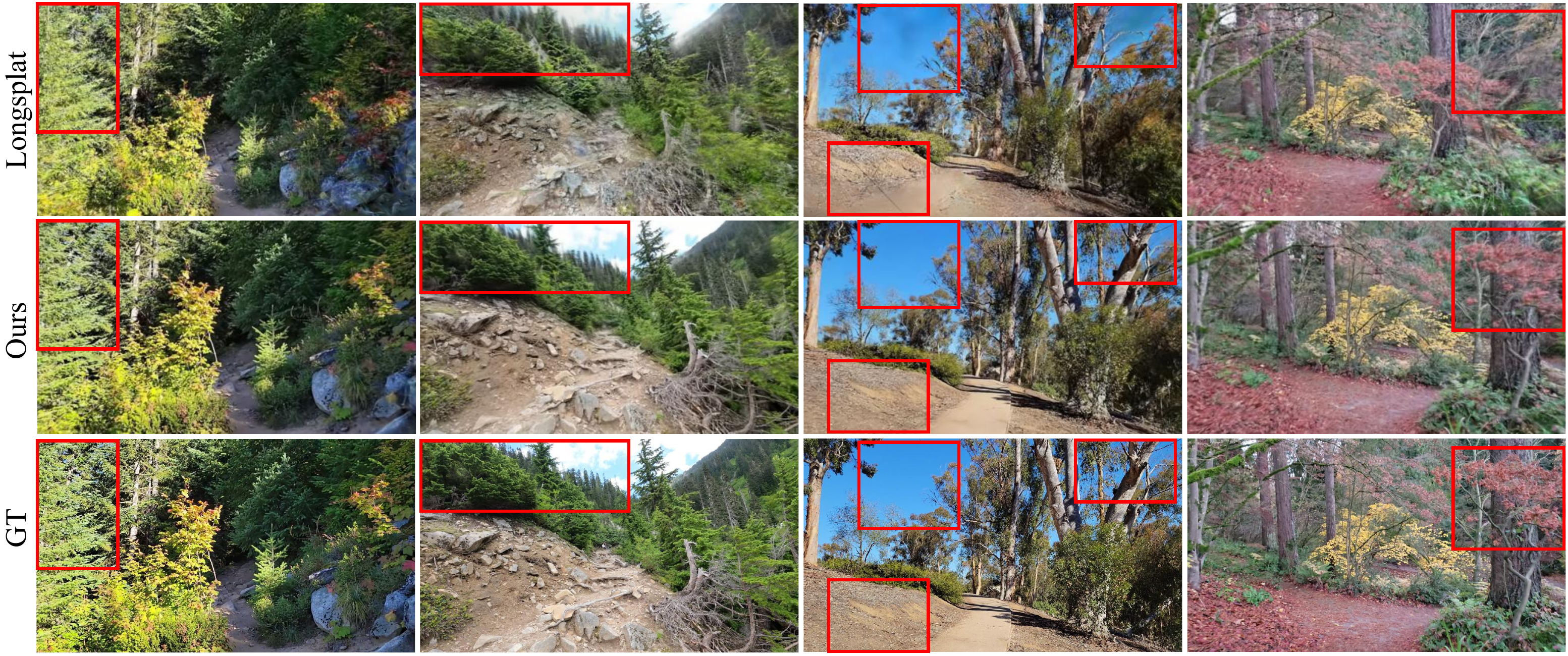}
\caption{\textbf{Qualitative results on the Hike dataset~\cite{hike}.} Compared with LongSplat~\cite{longsplat}, our method produces cleaner and more stable reconstructions across challenging outdoor scenes. In regions containing complex structures such as dense vegetation, LongSplat can exhibit blurring or produce unintended Gaussian artifacts in certain cases. In contrast, our method handles these situations more robustly overall and preserves more structurally consistent details.}
\label{fig:supp-fig17}
\end{figure*}

The qualitative results in Fig.~\ref{fig:supp-fig17} further support these findings. While reconstructions obtained with LongSplat often exhibit blurry regions or unintended Gaussian artifacts in structurally complex areas such as dense vegetation, our method reconstructs much clearer and more stable geometry. Overall, our proposed approach demonstrates stronger robustness to compression-related degradation and preserves more consistent structural details even in such challenging datasets.

\begin{table*}[t]
\centering
\scriptsize

\caption{\textbf{Quantitative evaluation on the Hike dataset \cite{hike}.}
Our method consistently outperforms the LongSplat baseline across diverse scenes in terms of both photometric quality and pose accuracy. The results demonstrate improved robustness under long trajectories and challenging camera motions.}
\label{tab:supp-tab21}

\resizebox{\textwidth}{!}{%
\begin{tabular}{l|cccccc|cccccc}
\toprule
\multirow{2}{*}{\textbf{Scene}} &
\multicolumn{6}{c|}{\textbf{LongSplat~\cite{longsplat}}} &
\multicolumn{6}{c}{\textbf{LongSplat + Ours}} \\
\cmidrule(lr){2-7} \cmidrule(lr){8-13}
& PSNR$\uparrow$ & SSIM$\uparrow$ & LPIPS$\downarrow$ & RPE$_t\downarrow$ & RPE$_r\downarrow$ & ATE$\downarrow$
& PSNR$\uparrow$ & SSIM$\uparrow$ & LPIPS$\downarrow$ & RPE$_t\downarrow$ & RPE$_r\downarrow$ & ATE$\downarrow$ \\
\midrule

forest1
& 18.78 & 0.45 & 0.37 & \textbf{0.447} & 5.938 & \textbf{0.009}
& \textbf{19.14} & \textbf{0.48} & \textbf{0.36} & 0.457 & \textbf{5.513} & \textbf{0.009} \\

forest2
& 24.96 & 0.76 & 0.20 & 0.421 & 0.908 & 0.012
& \textbf{25.98} & \textbf{0.81} & \textbf{0.17} & \textbf{0.332} & \textbf{0.558} & \textbf{0.011} \\

forest3
& 15.00 & 0.38 & \textbf{0.45} & 1.560 & 12.286 & \textbf{0.067}
& \textbf{15.11} & \textbf{0.39} & \textbf{0.45} & \textbf{1.551} & \textbf{11.158} & \textbf{0.067} \\

garden1
& 23.32 & \textbf{0.73} & \textbf{0.20} & \textbf{0.375} & 0.999 & \textbf{0.010}
& \textbf{23.37} & \textbf{0.73} & \textbf{0.20} & 0.492 & \textbf{0.944} & 0.011 \\

garden2
& 17.57 & 0.36 & 0.46 & \textbf{2.287} & 11.308 & 0.034
& \textbf{17.67} & \textbf{0.37} & \textbf{0.45} & 2.320 & \textbf{10.880} & \textbf{0.033} \\

garden3
& 19.42 & 0.43 & 0.41 & 1.169 & 6.450 & \textbf{0.015}
& \textbf{19.96} & \textbf{0.47} & \textbf{0.38} & \textbf{0.846} & \textbf{5.453} & \textbf{0.015} \\

indoor
& \textbf{17.74} & \textbf{0.63} & \textbf{0.42} & 0.959 & 9.635 & 0.033
& 17.27 & 0.62 & \textbf{0.42} & \textbf{0.761} & \textbf{7.311} & \textbf{0.031} \\

playground
& 17.82 & 0.42 & 0.42 & \textbf{1.994} & \textbf{8.748} & \textbf{0.031}
& \textbf{18.15} & \textbf{0.44} & \textbf{0.40} & 2.079 & 8.952 & \textbf{0.031} \\

university1
& \textbf{19.22} & \textbf{0.48} & \textbf{0.41} & \textbf{1.485} & \textbf{4.852} & 0.029
& 19.12 & 0.47 & 0.43 & 1.554 & 5.068 & \textbf{0.023} \\

university2
& 25.38 & 0.74 & 0.20 & 0.895 & 3.833 & 0.012
& \textbf{25.55} & \textbf{0.76} & \textbf{0.19} & \textbf{0.744} & \textbf{3.374} & \textbf{0.011} \\

university3
& 18.23 & 0.49 & 0.38 & 1.545 & \textbf{7.978} & 0.017
& \textbf{19.05} & \textbf{0.53} & \textbf{0.35} & \textbf{1.462} & 8.415 & \textbf{0.015} \\

university4
& 14.95 & 0.34 & 0.45 & \textbf{1.020} & 6.733 & \textbf{0.027}
& \textbf{15.46} & \textbf{0.37} & \textbf{0.44} & 1.226 & \textbf{6.594} & \textbf{0.027} \\

\midrule
Avg.
& 19.37 & 0.52 & 0.36 & 1.180 & 6.639 & 0.025
& \textbf{19.65} & \textbf{0.54} & \textbf{0.35} & \textbf{1.152} & \textbf{6.185} & \textbf{0.024} \\
\bottomrule
\end{tabular}%
}
\end{table*}

